%% file: main.tex
\documentclass[10pt,twocolumn,letterpaper]{article}

\usepackage{cvpr}              


\usepackage{colortbl}
\usepackage{booktabs}
\usepackage{algorithm}
\usepackage{algorithmic}

\newcommand{\cbox}[1]{\textcolor{#1}{\rule{1ex}{1ex}}}

\definecolor{road}{RGB}{128,64,128}
\definecolor{sidewalk}{RGB}{244,35,232}
\definecolor{building}{RGB}{70,70,70}
\definecolor{wall}{RGB}{102,102,156}
\definecolor{fence}{RGB}{190,153,153}
\definecolor{pole}{RGB}{153,153,153}
\definecolor{trafficlight}{RGB}{250,170,30}
\definecolor{trafficsign}{RGB}{220,220,0}
\definecolor{vegetation}{RGB}{107,142,35}
\definecolor{terrain}{RGB}{152,251,152}
\definecolor{pedestrian}{RGB}{220,20,60}
\definecolor{rider}{RGB}{255,0,0}
\definecolor{car}{RGB}{0,0,142}
\definecolor{truck}{RGB}{0,0,70}
\definecolor{bus}{RGB}{0,60,100}
\definecolor{motorcycle}{RGB}{0,0,230}
\definecolor{bicycle}{RGB}{119,11,32}
\definecolor{static}{RGB}{110,190,160}
\definecolor{dynamic}{RGB}{170,120,50}
\definecolor{other}{RGB}{55,90,80}
\definecolor{water}{RGB}{45,60,150}
\definecolor{roadline}{RGB}{157,234,50}
\definecolor{ground}{RGB}{81,0,81}
\definecolor{bridge}{RGB}{150,100,100}
\definecolor{railtrack}{RGB}{230,150,140}
\definecolor{guardrail}{RGB}{180,165,180}
\definecolor{cvprblue}{rgb}{0.21,0.49,0.74}
\definecolor{lightblue}{RGB}{230,240,250}
\definecolor{lightgray}{RGB}{230,230,230}
\usepackage[pagebackref,breaklinks,colorlinks,allcolors=cvprblue]{hyperref}

\usepackage{multirow}

\def\confName{CVPR}
\def\confYear{2027}

\title{AerialOcc: Unified 3D Aerial Perception without Task-Specific Head}

\author{
\textbf{Fan Liu}\textsuperscript{1,2
},
\textbf{Xiang Gu}\textsuperscript{1},
\textbf{Baoyu Lv}\textsuperscript{1},
\textbf{Chuanyi Zhang}\textsuperscript{1},
\textbf{Liang Yao}\textsuperscript{1},
\textbf{Yuye Ma}\textsuperscript{1},
\textbf{Yuhui Zheng}\textsuperscript{1}
\\
\textsuperscript{1}Hohai University
\\
\textsuperscript{2}Key Laboratory of Water Big Data Technology of Ministry of Water Resources
\\
{\tt\small Email: fanliu@hhu.edu.cn}
}

\begin{document}
\maketitle
\input{sec/0_abstract}    
\input{sec/1_intro}
\input{sec/2_related}
\input{sec/3_dataset}

\input{sec/4_aerial}
\input{sec/5_experiments}
\input{sec/6_conclusion}

{
    \small
    \bibliographystyle{ieeenat_fullname}
    \bibliography{main}
}


\clearpage

\onecolumn

\begin{center}
    {\LARGE\bfseries Supplementary Material}
\end{center}

\vspace{0.5cm}

\input{sec/X_suppl}

\end{document}


\maketitle
\thispagestyle{empty}

\section*{A. More Dataset Composition Details}

AerialOcc-20K comprises 20 scenes generated across four CARLA maps: Town03, Town05, Town07, and Town10HD. Table~\ref{tab:supp_scene_composition} presents their allocation across the training and validation splits. The benchmark contains 16,000 training samples and 4,000 validation samples, without a separate test split.

All 20,000 samples pass the synchronization integrity check, with complete RGB observations and measurements from all 24 semantic LiDAR sensors, and no missing sensor frames or out-of-range semantic labels.Detailed scene-level composition is provided in Table~\ref{tab:supp_scene_composition}.

\begin{table*}[ht]
\centering
\setlength{\tabcolsep}{10pt}
\begin{tabular}{@{}lccc@{}}
\toprule
Town & Scene IDs & \# Scenes & \# Samples \\
\midrule
\multicolumn{4}{@{}l}{\textbf{Training}} \\
Town03   & 01, 04, 07 & 3 & 3,000 \\
Town05   & 00, 03, 06 & 3 & 3,000 \\
Town07   & 02, 05, 08 & 3 & 3,000 \\
Town10HD & 09--15     & 7 & 7,000 \\
\midrule
\multicolumn{4}{@{}l}{\textbf{Validation}} \\
Town03   & 17 & 1 & 1,000 \\
Town05   & 16 & 1 & 1,000 \\
Town07   & 18 & 1 & 1,000 \\
Town10HD & 19 & 1 & 1,000 \\
\midrule
\textbf{Total} & 00--19 & \textbf{20} & \textbf{20,000} \\
\bottomrule
\end{tabular}
\caption{Scene-level training and validation composition of AerialOcc-20K.}
\label{tab:supp_scene_composition}
\end{table*}

\begin{table}[t]
\centering
\setlength{\tabcolsep}{2.5pt}
\begin{tabular*}{\columnwidth}{@{\extracolsep{\fill}}c|c|c|c@{}}
\toprule
\textbf{Scene} & \textbf{Town} & \textbf{Scenario} & \textbf{Samples} \\
\midrule
\multicolumn{1}{@{}c}{\textbf{Training}} & \multicolumn{2}{c}{} & \multicolumn{1}{c@{}}{\textbf{16,000}} \\
scene\_00 & Town05 & Urban Grid & 1,000 \\
scene\_01 & Town03 & Urban Junction & 1,000 \\
scene\_02 & Town07 & Rural & 1,000 \\
scene\_03 & Town05 & Urban Grid & 1,000 \\
scene\_04 & Town03 & Urban Junction & 1,000 \\
scene\_05 & Town07 & Rural & 1,000 \\
scene\_06 & Town05 & Urban Grid & 1,000 \\
scene\_07 & Town03 & Urban Junction & 1,000 \\
scene\_08 & Town07 & Rural & 1,000 \\
scene\_09 & Town10HD & Urban Downtown & 1,000 \\
scene\_10 & Town10HD & Urban Downtown & 1,000 \\
scene\_11 & Town10HD & Urban Downtown & 1,000 \\
scene\_12 & Town10HD & Urban Downtown & 1,000 \\
scene\_13 & Town10HD & Urban Downtown & 1,000 \\
scene\_14 & Town10HD & Urban Downtown & 1,000 \\
scene\_15 & Town10HD & Urban Downtown & 1,000 \\
\addlinespace[2pt]
\midrule
\multicolumn{1}{@{}c}{\textbf{Validation}} & \multicolumn{2}{c}{} & \multicolumn{1}{c@{}}{\textbf{4,000}} \\
scene\_16 & Town05 & Urban Grid & 1,000 \\
scene\_17 & Town03 & Urban Junction & 1,000 \\
scene\_18 & Town07 & Rural & 1,000 \\
scene\_19 & Town10HD & Urban Downtown & 1,000 \\
\addlinespace[2pt]
\midrule
\multicolumn{1}{@{}c}{\textbf{Total}} & \multicolumn{2}{c}{} & \multicolumn{1}{c@{}}{\textbf{20,000}} \\
\bottomrule
\end{tabular*}
\caption{Statistics of AerialOcc-20K. The dataset comprises 20 scenes across four CARLA towns, with distinct collection trajectories for each scene.}
\label{tab:aerialocc20k_statistics}
\end{table}

\section*{B. More Method Details}
This section provides additional details of the training and inference procedures of AerialOcc. Algorithm~\ref{alg:supp_training} summarizes the joint optimization of geometry-aware occupancy learning and instance-aware supervision. Algorithm~\ref{alg:supp_inference} presents the unified inference procedure, in which semantic scene completion, 3D semantic segmentation, and 3D object detection are derived from the same semantic occupancy prediction.
\begin{algorithm}[h]
\caption{Training Procedure of AerialOcc}
\label{alg:supp_training}
\begin{algorithmic}[1]
\REQUIRE Current and historical images \(\mathcal{I}_t\), camera intrinsics \(\mathcal{K}_t\), relative poses \(\mathcal{T}_t\), voxel annotations \(Y_t\), valid observation mask \(m_t(v)\), object labels \(C_t^{*}\), and object boxes \(B_t^{*}\)
\FOR{each training iteration}
\STATE Extract multi-scale features from the current and historical images
\STATE Transform historical features into the current coordinate frame
\STATE Aggregate historical and current features into \(F_t^{\mathrm{temp}}\)
\STATE Estimate the depth distribution and construct \(D_t\)
\STATE Apply KL-based distillation from VGGT pseudo-depth soft labels
\STATE Construct voxel features \(V_t=\mathcal{E}_{\mathrm{occ}}(Q_v,F_t^{\mathrm{temp}},D_t)\)
\STATE Predict occupancy logits \(O_t=H_{\mathrm{occ}}(V_t)\)
\STATE Compute \(P_t=\operatorname{Softmax}(O_t)\)
\STATE Obtain \(P_t^{\mathrm{occ}}\) and \(Y_t^{\mathrm{occ}}\)
\STATE Compute \(\mathcal{L}_{\mathrm{occ}}\) over voxels satisfying \(m_t(v)=1\)
\STATE Select and pool object-relevant features from \(V_t\) into \(F_t^{\mathrm{obj}}\)
\STATE Decode learnable object queries into object features \(Z_t\)
\STATE Predict \(\hat{C}_t^{\mathrm{aux}}\) and \(\hat{B}_t^{\mathrm{aux}}\)
\STATE Compute \(\mathcal{L}_{\mathrm{det}}\)
\STATE Construct \(M_t^{\mathrm{box}}\) and compute \(\mathcal{L}_{\mathrm{box\text{-}occ}}\)
\STATE Compute \(\mathcal{L}_{\mathrm{joint}}\)
\STATE Update the trainable parameters using AdamW
\ENDFOR
\end{algorithmic}
\end{algorithm}
\begin{algorithm}[h]
\caption{Unified Inference Procedure of AerialOcc}
\label{alg:supp_inference}
\begin{algorithmic}[1]
\REQUIRE Current and historical images \(\mathcal{I}_t\), camera intrinsics \(\mathcal{K}_t\), and relative poses \(\mathcal{T}_t\)
\ENSURE Semantic scene completion, 3D semantic segmentation, and 3D object detection results
\STATE Extract multi-scale features from the current and historical images
\STATE Transform historical features into the current coordinate frame
\STATE Aggregate historical and current features into \(F_t^{\mathrm{temp}}\)
\STATE Estimate the depth distribution and construct \(D_t\)
\STATE Construct voxel features \(V_t=\mathcal{E}_{\mathrm{occ}}(Q_v,F_t^{\mathrm{temp}},D_t)\)
\STATE Predict occupancy logits \(O_t=H_{\mathrm{occ}}(V_t)\)
\STATE Compute \(P_t=\operatorname{Softmax}(O_t)\)
\STATE Obtain semantic scene completion using \(\Phi_{\mathrm{ssc}}(O_t)\)
\STATE Assign semantic labels using \(\hat{S}_t(v)=\mathop{\arg\max}_{c\in\mathcal{C}}P_t(c\mid v)\)
\FOR{each \(c\in\mathcal{C}_{\mathrm{obj}}\)}
\STATE Construct the class-specific soft mask \(S_t^c(v)=P_t(c\mid v)\)
\STATE Apply EPOC to separate individual instance regions
\STATE Apply M2B to convert the separated regions into oriented 3D bounding boxes
\STATE Add the predicted boxes to \(\hat{B}_t\)
\ENDFOR
\STATE \textbf{return} \(\Phi_{\mathrm{ssc}}(O_t)\), \(\hat{S}_t(v)\), and \(\hat{B}_t\)
\end{algorithmic}
\end{algorithm}

\section*{C. More Implementation Details}
We adopt EfficientNet-B7 as the shared image backbone. AerialOcc is optimized for 25,000 iterations using AdamW with a maximum learning rate of \(3\times10^{-4}\), weight decay of \(0.01\). We employ OneCycleLR with cosine annealing, where the first \(5\%\) of the optimization process, approximately 1,250 iterations, is used for learning-rate warm-up. The batch size is set to 1 per GPU, resulting in a total batch size of 2 on two NVIDIA GeForce RTX 4090 GPUs. All experiments are conducted in FP32 without mixed-precision training, gradient clipping, or gradient accumulation.
The input images are resized to \(704\times384\). The model-side occupancy range is \([-25.6,51.2]\times[-25.6,25.6]\times[-32,8]\,\mathrm{m}\). The latent voxel grid has a resolution of \(128\times128\times16\), which is decoded into an output grid of \(256\times256\times32\).

\clearpage
\section*{D. More Experiments}
\subsection*{D.1 Per-Class Occupancy Results}
Table~\ref{tab:occ_perclass} reports the per-class semantic occupancy results of AerialOcc on AerialOcc-20K. AerialOcc achieves an overall IoU of 70.29\% and an mIoU of 28.13\%. It performs particularly well on major structural categories, including road, sidewalk, terrain, water, and ground, with IoUs above 70\%. Strong results are also obtained for rail track, car, building, and wall. In contrast, rare or small categories such as traffic light, pedestrian, truck, and guard rail remain challenging, while bus and bridge receive zero IoU. These results indicate that class imbalance and limited observations of small objects remain the main difficulties in fine-grained aerial occupancy prediction.

\begin{table*}[h]
\centering
\setlength{\tabcolsep}{12pt}
\begin{tabular}{lcc}
\toprule
\textbf{Category}
& \textbf{Distribution}
& \textbf{AerialOcc} \\
\midrule
\cbox{road} Road
& 30.13
& 87.76 \\
\cbox{sidewalk} Sidewalk
& 18.03
& 84.39 \\
\cbox{building} Building
& 22.57
& 42.03 \\
\cbox{wall} Wall
& 0.50
& 41.32 \\
\cbox{fence} Fence
& 0.34
& 17.60 \\
\cbox{pole} Pole
& 2.09
& 12.72 \\
\cbox{trafficlight} Traffic light
& 0.01
& 4.92 \\
\cbox{trafficsign} Traffic sign
& 0.04
& 10.67 \\
\cbox{vegetation} Vegetation
& 9.03
& 26.53 \\
\cbox{terrain} Terrain
& 8.20
& 73.28 \\
\cbox{pedestrian} Pedestrian
& 0.02
& 5.11 \\
\cbox{rider} Rider
& 0.01
& 14.56 \\
\cbox{car} Car
& 1.27
& 42.45 \\
\cbox{truck} Truck
& 0.52
& 5.25 \\
\cbox{bus} Bus
& 0.17
& 0.00 \\
\cbox{motorcycle} Motorcycle
& 0.01
& 14.28 \\
\cbox{bicycle} Bicycle
& 0.00
& 10.65 \\
\cbox{static} Static
& 0.38
& 20.64 \\
\cbox{dynamic} Dynamic
& 0.49
& 33.53 \\
\cbox{other} Other
& 0.34
& 23.42 \\
\cbox{water} Water
& 0.31
& 74.38 \\
\cbox{roadline} Road line
& 1.04
& 18.92 \\
\cbox{ground} Ground
& 0.55
& 70.29 \\
\cbox{bridge} Bridge
& 0.00
& 0.00 \\
\cbox{railtrack} Rail track
& 3.87
& 50.87 \\
\cbox{guardrail} Guard rail
& 0.04
& 1.81 \\
\midrule
\multicolumn{2}{l}{\textbf{Overall IoU}}
& \textbf{70.29} \\
\multicolumn{2}{l}{\textbf{mIoU}}
& \textbf{28.13} \\
\bottomrule
\end{tabular}
\caption{Per-class semantic scene completion results of AerialOcc on the AerialOcc-20K validation set.}
\label{tab:occ_perclass}
\end{table*}

\clearpage
\subsection*{D.2 Collaborative Occupancy Prediction}
Table~\ref{tab:occupancy_co_results} compares AerialOcc with existing methods under collaborative and non-collaborative settings. Without collaboration, AerialOcc achieves 19.25\% mIoU, outperforming BEVDet and PanoOcc by 2.00 and 3.21 percentage points, respectively. With collaborative perception, AerialOcc reaches 27.31\% mIoU, exceeding MCOP by 4.02 percentage points. Collaboration improves the performance of AerialOcc by 8.06 percentage points, demonstrating that its occupancy representation can effectively incorporate complementary observations from multiple UAVs.

\begin{table*}[h]
\centering
\setlength{\tabcolsep}{18pt}
\begin{tabular}{lcc}
\toprule
\textbf{Method}
& \textbf{Co.}
& \textbf{mIoU}$\uparrow$ \\
\midrule
BEVDet\(^{\dagger}\)
& \(\times\)
& \underline{17.25} \\
PanoOcc\(^{\dagger}\)
& \(\times\)
& 16.04 \\
AerialOcc (Ours)
& \(\times\)
& \textbf{19.25} \\
\midrule
BEVDet\(^{\ddagger}\)
& \(\checkmark\)
& 22.34 \\
PanoOcc\(^{\ddagger}\)
& \(\checkmark\)
& 18.05 \\
MCOP
& \(\checkmark\)
& \underline{23.29} \\
AerialOcc (Ours)
& \(\checkmark\)
& \textbf{27.31} \\
\bottomrule
\end{tabular}
\caption{Semantic scene completion results with and without collaborative perception. Co. denotes collaboration, while \(^{\dagger}\) and \(^{\ddagger}\) indicate FlashOcc-based conversion and Where2comm-based collaboration, respectively.}
\label{tab:occupancy_co_results}
\end{table*}

\clearpage
\subsection*{D.3 Per-Class Results on OccuFly}
Table~\ref{tab:occufly_perclass} provides detailed results on OccuFly at different flight altitudes. Compared with CGFormer, AerialOcc improves occupancy IoU by 16.25, 17.08, and 23.82 percentage points at \(30\,\mathrm{m}\), \(40\,\mathrm{m}\), and \(50\,\mathrm{m}\), respectively. The corresponding mIoU improvements are 6.08, 2.54, and 6.31 percentage points. When all altitudes are combined, AerialOcc achieves 44.49\% IoU and 5.49\% mIoU, outperforming CGFormer by 19.83 and 4.69 percentage points. With only 10\% of the target-domain data for fine-tuning, AerialOcc obtains 5.44\% mIoU, only 0.21 percentage points below full-data fine-tuning. These results demonstrate consistent performance across altitude changes and effective transfer of semantic knowledge to real aerial environments.
\begin{table}[h]
\centering
\renewcommand{\arraystretch}{1.2}
\resizebox{\textwidth}{!}{%
\begin{tabular}{l|cc|cc|cc|cc|cc}
\hline
\multirow{2}{*}{\textbf{Category}}
& \multicolumn{2}{c|}{\textbf{30 m}}
& \multicolumn{2}{c|}{\textbf{40 m}}
& \multicolumn{2}{c|}{\textbf{50 m}}
& \multicolumn{2}{c|}{\textbf{All}}
& \multicolumn{2}{c}{\textbf{Sim-to-Real}} \\
\cline{2-11}
& \textbf{CGFormer}
& \textbf{AerialOcc}
& \textbf{CGFormer}
& \textbf{AerialOcc}
& \textbf{CGFormer}
& \textbf{AerialOcc}
& \textbf{CGFormer}
& \textbf{AerialOcc}
& \shortstack{\textbf{10\% FT}}
& \shortstack{\textbf{100\% FT}} \\
\hline
Road
& 3.87 & 11.10
& 2.21 & 2.42
& 0.28 & \textbf{20.09}
& 1.43 & 11.92
& 9.79 & \underline{12.23} \\
Walkway
& 1.22 & 2.07
& 0.55 & 0.17
& 0.55 & \textbf{6.39}
& 0.61 & 3.40
& 3.70 & \underline{3.82} \\
Dirt
& 0.17 & 1.24
& 0.09 & 0.22
& 0.31 & \textbf{4.23}
& 0.24 & 2.20
& \underline{2.67} & 2.32 \\
Gravel
& 0.16 & 1.20
& 0.23 & 0.30
& 0.44 & \underline{2.85}
& 0.35 & 1.71
& \textbf{2.97} & 2.09 \\
Rock
& 0.01 & 0.03
& 0.02 & 0.01
& \textbf{0.05} & 0.03
& \underline{0.04} & 0.02
& 0.01 & 0.03 \\
Grass
& 3.46 & \textbf{17.86}
& 1.03 & 3.81
& 8.51 & \underline{17.81}
& 5.35 & 11.53
& 15.36 & 14.58 \\
Vegetation
& 0.47 & \textbf{10.78}
& 0.35 & 2.62
& 0.92 & 4.18
& 0.76 & 4.07
& 4.30 & \underline{4.78} \\
Tree
& 0.54 & \textbf{20.12}
& 0.24 & 9.41
& 0.28 & 8.89
& 0.28 & 10.69
& \underline{11.78} & 10.35 \\
Ground-Obs.
& 0.98 & 1.43
& 0.09 & 0.24
& 0.41 & \textbf{6.58}
& 0.29 & 3.45
& 3.08 & \underline{4.16} \\
Person
& 0.00 & 0.00
& 0.00 & 0.00
& 0.00 & 0.00
& 0.00 & 0.00
& 0.00 & 0.00 \\
Bicycle
& 0.00 & 0.00
& 0.00 & 0.00
& 0.00 & 0.00
& 0.00 & 0.00
& \underline{0.03} & \textbf{0.04} \\
Vehicle
& 0.25 & \underline{14.21}
& 0.18 & 4.27
& 0.02 & \textbf{16.96}
& 0.11 & 10.99
& 13.09 & 11.66 \\
Water
& 0.00 & 0.00
& 0.00 & 0.00
& 0.00 & 0.00
& 0.00 & 0.00
& 0.00 & 0.00 \\
Building
& 9.59 & \underline{42.47}
& 7.67 & 34.89
& 7.90 & \textbf{46.37}
& 7.88 & 41.30
& 31.26 & 38.18 \\
Roof
& 0.68 & \textbf{10.52}
& 0.17 & 2.87
& 0.16 & \underline{7.83}
& 0.19 & 5.94
& 6.17 & 6.16 \\
Cables
& 0.00 & 0.00
& 0.00 & 0.00
& 0.00 & 0.00
& 0.00 & 0.00
& \textbf{0.01} & 0.00 \\
Cable-Tower
& 0.00 & 0.00
& 0.00 & 0.00
& 0.00 & 0.00
& 0.00 & 0.00
& 0.00 & 0.00 \\
Flying-Animals
& 0.00 & 0.00
& 0.00 & 0.00
& 0.00 & 0.00
& 0.00 & 0.00
& 0.00 & 0.00 \\
Parking-Lot
& 0.01 & \textbf{17.01}
& 0.00 & 5.29
& 0.01 & \underline{12.99}
& 0.01 & 10.22
& 10.51 & 9.49 \\
Constructions
& 0.05 & 0.71
& 0.06 & 0.04
& 0.15 & \textbf{0.98}
& 0.13 & 0.59
& 0.29 & \underline{0.78} \\
Cranes
& 0.00 & 0.00
& 0.00 & 0.00
& 0.00 & 0.00
& 0.00 & 0.00
& 0.00 & 0.00 \\
Truck
& 0.00 & \underline{4.28}
& 0.00 & 2.26
& 0.00 & 2.62
& 0.00 & 2.84
& \textbf{4.76} & 3.72 \\
\hline
\textbf{mIoU}
& 0.97 & \underline{7.05}
& 0.59 & 3.13
& 0.91 & \textbf{7.22}
& 0.80 & 5.49
& 5.44 & 5.65 \\
\textbf{IoU}
& 29.48 & \underline{45.73}
& 20.82 & 37.90
& 26.00 & \textbf{49.82}
& 24.66 & 44.49
& 33.39 & 41.19 \\
\hline
\end{tabular}%
}
\caption{Per-class semantic scene completion IoU results on the OccuFly validation set at different flight altitudes and adaptation settings. All results are reported in percentage form. FT denotes fine-tuning.}
\label{tab:occufly_perclass}
\end{table}

\clearpage
\section*{E. More Qualitative Analysis}
\subsection*{E.1 Dataset}

\begin{figure*}[h]
\centering
\includegraphics[width=0.78\textwidth, height=0.72\textheight, keepaspectratio=false]{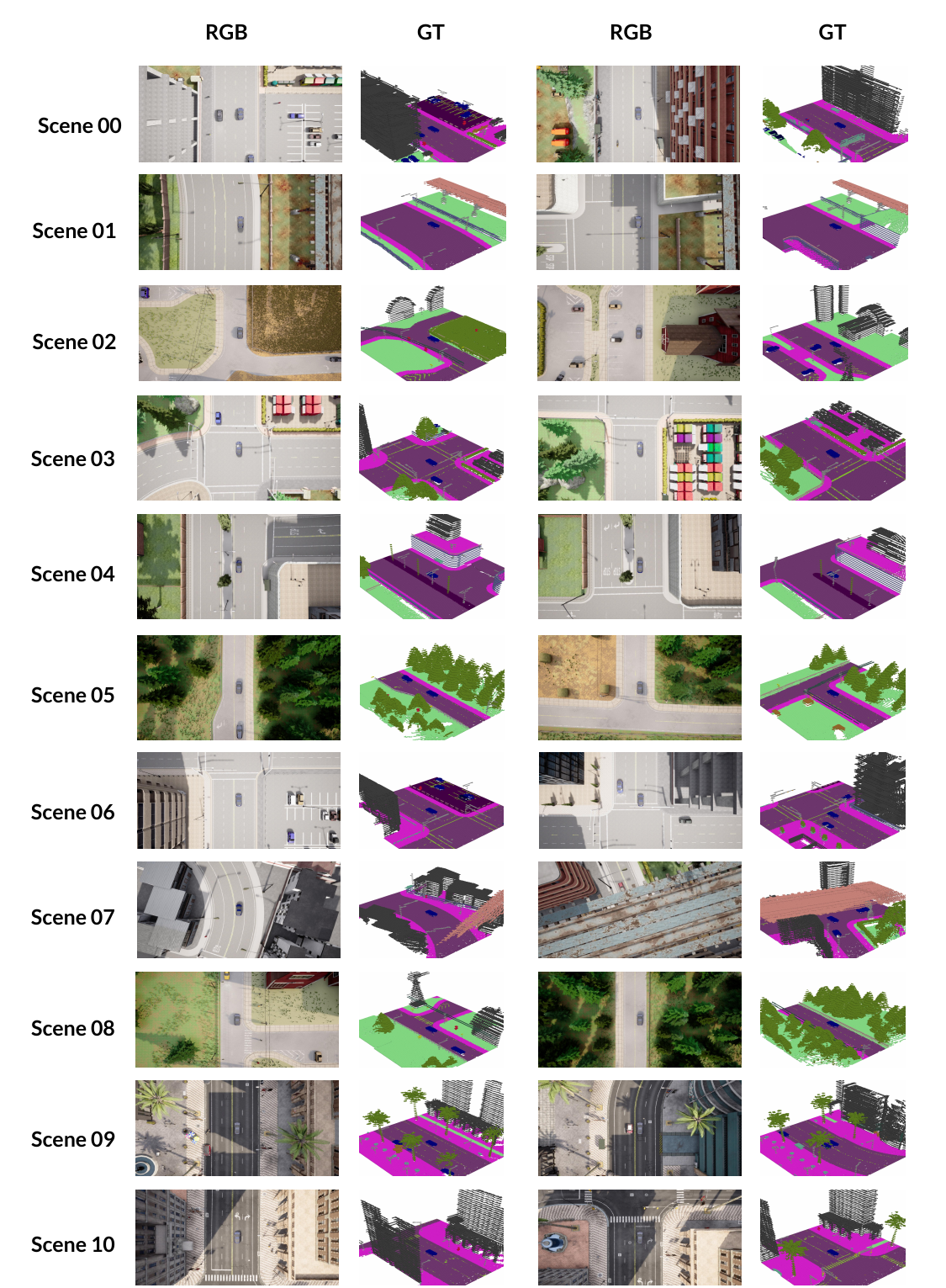}
\caption{Qualitative examples of RGB observations and the corresponding 3D semantic occupancy ground-truth annotations from scenes 00--10 in AerialOcc-20K.}
\label{fig:SupplementaryGT1}
\end{figure*}

\begin{figure*}[h]
\centering
\includegraphics[width=0.78\textwidth, height=0.72\textheight, keepaspectratio=false]{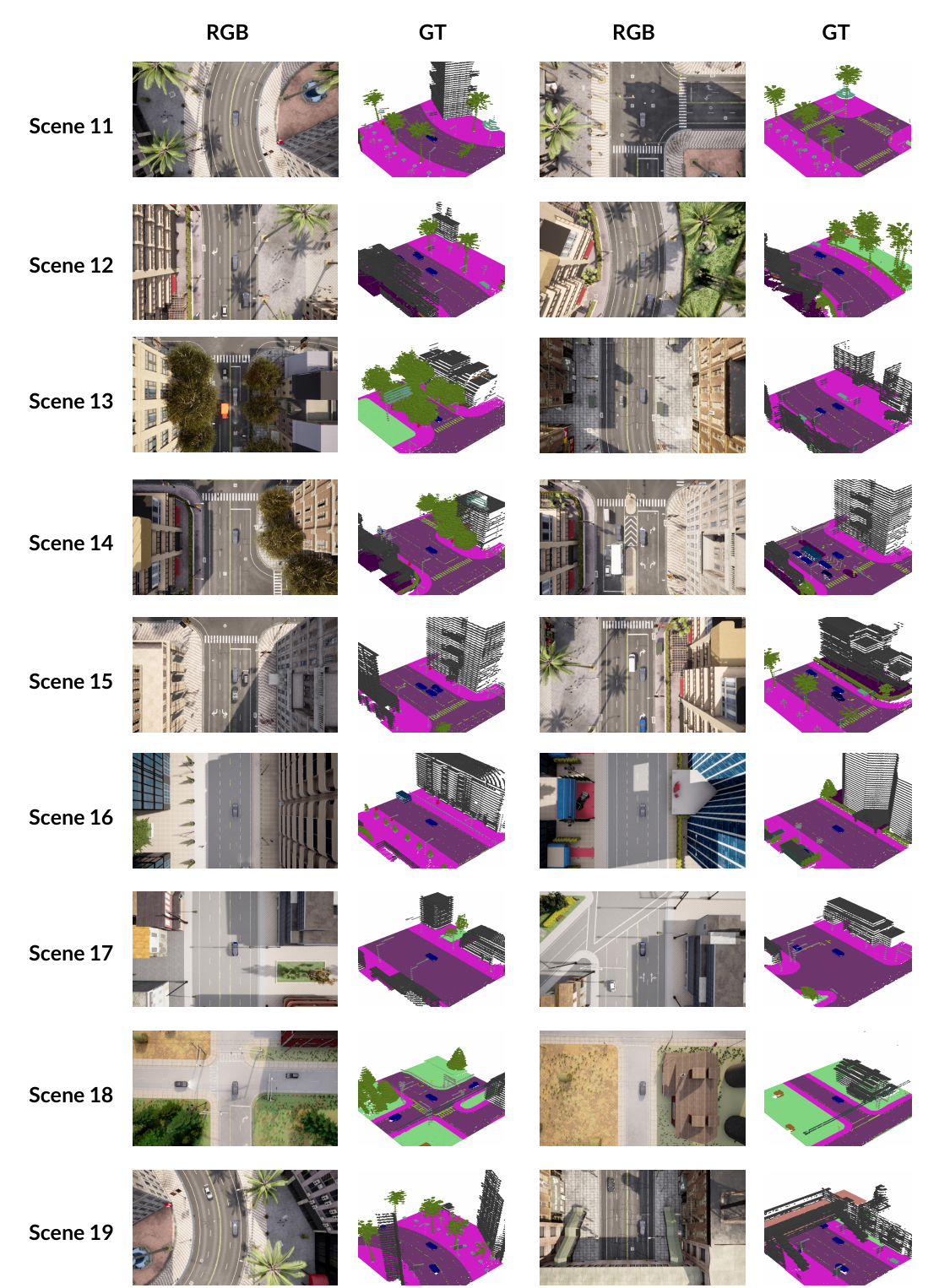}
\caption{Qualitative examples of RGB observations and the corresponding 3D semantic occupancy ground-truth annotations from scenes 11--19 in AerialOcc-20K.}
\label{fig:SupplementaryGT2}
\end{figure*}

\clearpage
\subsection*{E.2 Qualitative results of AerialOcc}

\begin{figure*}[h]
\centering
\includegraphics[width=1\textwidth]{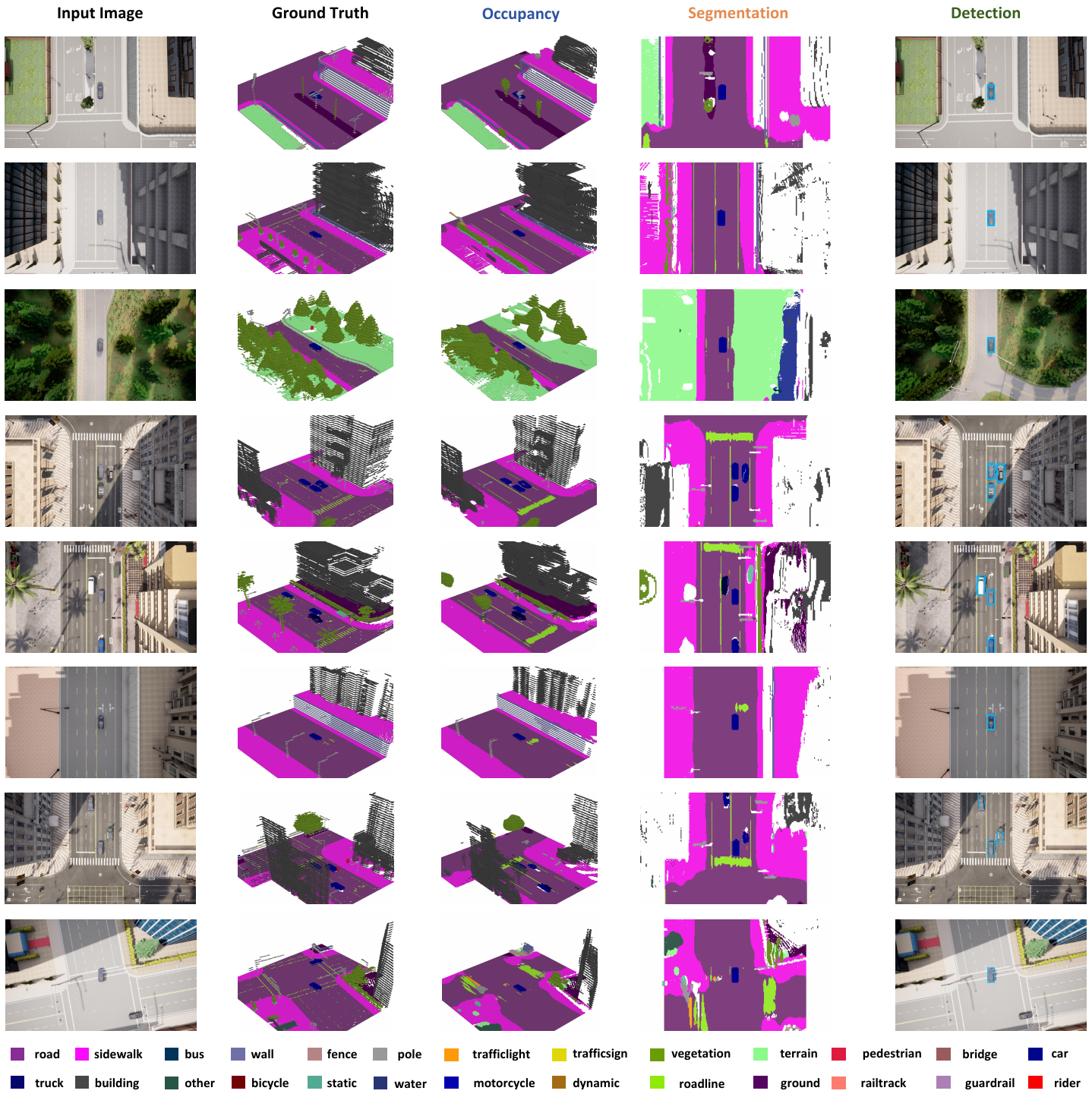} 
\caption{More qualitative results of AerialOcc on semantic scene completion, 3D voxel semantic segmentation, and 3D object detection.}
\label{fig:SupplementaryVisualization}
\end{figure*}


\clearpage
\section*{F. Limitations and Future Work}
AerialOcc currently adopts a fixed-resolution dense voxel grid to provide a consistent representation across different perception tasks. Although this design facilitates unified prediction and task conversion, its computational cost increases with the spatial range and voxel resolution. Existing sparse, hierarchical, or adaptive voxel representations could be incorporated to improve efficiency without changing the occupancy-centric formulation.

The current framework focuses on semantic scene completion, 3D voxel semantic segmentation, and 3D object detection. Future work may enrich the occupancy representation with additional attributes, such as motion, uncertainty, or temporal association, to support tasks including tracking and forecasting. These extensions are compatible with the unified occupancy interface and do not require redesigning the core framework.

\section*{G. Ethical Statement}
AerialOcc-20K is generated entirely in CARLA and therefore does not contain real-world personal information or identifiable individuals. The real-world benchmarks used in our experiments are employed only for research evaluation under their respective licenses.

Aerial perception technologies may support beneficial applications such as inspection, environmental monitoring, and autonomous navigation, but could also be misused for intrusive surveillance or unsafe autonomous operation. Practical deployment should comply with applicable privacy and aviation regulations and include appropriate safety constraints, human oversight, and access control.

%% file: sec/0_abstract.tex
\begin{abstract}
Aerial perception systems are increasingly expected to support diverse tasks, yet existing solutions remain fundamentally tied to task-specific outputs. Detection, segmentation, and other perception objectives are usually learned either by separate models or by sharing a backbone with multiple task heads, leaving the model itself unaware of the common structure underlying these tasks.
We are interested in asking: can a single perception representation support different levels of understanding without being designed for any specific task? We notice that occupancy is a promising task-agnostic representation for drone perception. By modeling the spatial existence and semantics of the environment, it naturally bridges different perception granularities, from coarse scene understanding to fine-grained object localization. Therefore, we introduce AerialOcc, a unified occupancy-based perception model that enables multiple drone perception tasks from a single inference process, without maintaining task-specific prediction heads. To facilitate this direction, we construct AerialOcc-20K, a large-scale multi-class occupancy dataset for aerial environments. Extensive experiments show that AerialOcc achieves competitive performance across diverse perception tasks while preserving a unified architecture. Our findings demonstrate that occupancy can serve as a universal perception interface, offering a new perspective toward flexible and scalable visual intelligence for autonomous drones from a common 3D scene representation. Models and data are publicly available at \url{https://github.com/guxiang2002/AerialOcc}.  
\end{abstract}

%% file: sec/1_intro.tex
\section{Introduction}
\label{sec:intro}

Unmanned Aerial Vehicles (UAVs) play a vital role in applications such as autonomous navigation, infrastructure inspection, and environmental monitoring. To effectively handle such complex environments, UAV systems require comprehensive 3D scene understanding~\citep{gross2026occufly,ye2024uav3d}, spanning semantic scene completion, 3D semantic segmentation, and 3D object detection. However, these tasks describe different aspects of the same environment through heterogeneous output representations. Treating them independently makes it difficult to establish a consistent scene-level understanding, limiting the integration of complementary perception results and complicating downstream decision-making.

\begin{figure}[t]
\centering
\includegraphics[width=\columnwidth]{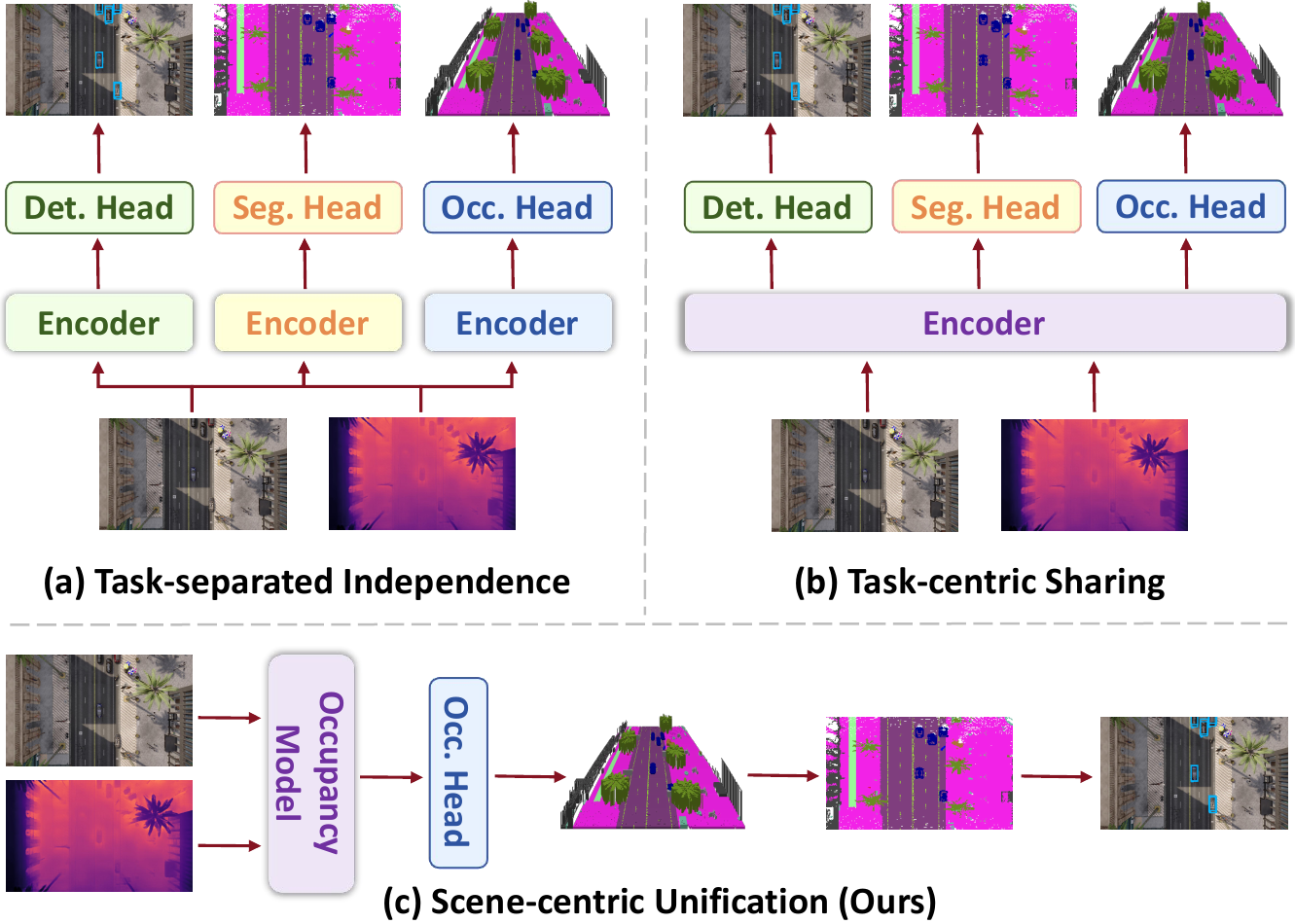}
\caption{Comparison of aerial perception paradigms. (a) Conventional task-separated perception learns independent representations for different tasks.  (b) Task-centric sharing improves efficiency by sharing an encoder but still relies on task-specific prediction heads.  (c) The proposed scene-centric formulation learns a unified occupancy representation that supports occupancy prediction, segmentation, and detection.}
\label{fig:Motivation}
\end{figure}

Current attempts to unify aerial perception tasks mainly fall into two paradigms, as illustrated in Fig.~\ref{fig:Motivation}. Task-separated methods~\citep{milioto2019rangenet++,yin2021center} train independent models for different objectives, resulting in duplicated feature extraction and isolated task representations. Task-centric methods share an encoder~\citep{ye2023lidarmultinet,chen2025m3net} to reduce redundant feature learning, but still rely on dedicated prediction heads tied to predefined outputs. Consequently, supporting new tasks requires additional task-specific components and supervision, restricting architectural flexibility and hindering the extension of these methods to broader perception scenarios.

These limitations motivate us to design a flexible aerial perception architecture that supports multiple tasks and output granularities. Semantic occupancy has been widely studied in autonomous driving~\citep{huang2023tri,li2023voxformer,yu2024context,wang2025uniocc} and recently extended to aerial scenes~\citep{lin2025mcop,gao2026skyshield,gross2026occufly}. However, existing methods primarily focus on occupancy prediction as an individual task, rather than exploiting it as a shared representation for multiple perception objectives. Our proposed model operates at the voxel level, the basic unit of a 3D scene perception. By assigning each voxel a spatial state and semantic category, the same representation can produce complete scene geometry, semantic regions, and object instances. This makes semantic occupancy a natural common interface for integrating geometric, semantic, and instance-level information within a unified 3D representation.

To equip such a unified representation with dense geometric and semantic understanding, voxel-wise supervision is required. However, existing UAV datasets mainly provide 2D boxes~\cite{jiang2021anti,rizzoli2023syndrone,khose2024skyscenes,yao2025uemm}, pixel-wise labels~\cite{azimi2019skyscapes,cai2025vdd,fonder2019mid,kolbeinsson2024ddos,marcu2023self,rizzoli2023syndrone,khose2024skyscenes}, depth maps~\cite{fonder2019mid,kolbeinsson2024ddos,marcu2023self,khose2024skyscenes}, or 3D boxes~\cite{ye2024uav3d,dhaouadi2025highlydata,lin2022capturing}. Although aerial occupancy datasets have emerged, existing resources remain limited in semantic coverage and annotation diversity. We therefore construct AerialOcc-20K in CARLA~\cite{dosovitskiy2017carla}, providing dense geometric and semantic annotations across diverse aerial environments, with the benefits of accurate labeling and reproducible data generation offered by simulation.

Building on this insight and dataset, we present AerialOcc, a unified aerial perception model built upon a novel architectural paradigm centered on semantic scene completion. Our model leverages the dense voxel-level outputs inherent in semantic occupancy, converting them into the formats required by other vision-centric tasks. AerialOcc supports scene-, region-, and instance-level perception within a unified architecture, without the need to design a separate task-specific head for each task. 


Extensive experiments demonstrate the effectiveness of AerialOcc across the three perception tasks. On AerialOcc-20K, AerialOcc achieves 69.02\% IoU and 29.93\% mIoU for semantic scene completion, substantially outperforming the compared occupancy baseline. It also surpasses task-specific methods in 3D semantic segmentation and 3D object detection. Furthermore, its strong performance on OccuFly and competitive results on UAV3D demonstrate the transferability of the learned occupancy representation to real-world aerial environments.

The contributions of this paper to aerial perception are summarized as follows:

\begin{itemize}

\item We propose AerialOcc, an occupancy-centric aerial perception model that derives semantic scene completion, 3D semantic segmentation, and 3D object detection through a unified inference pipeline based on shared occupancy representations, without other task-specific prediction heads.

\item To support the training and evaluation of AerialOcc, we construct AerialOcc-20K, a large-scale multi-class aerial occupancy dataset with dense voxel-wise semantic annotations across diverse altitudes and viewpoints for unified 3D perception.

\item Extensive experiments demonstrate strong performance across multiple perception tasks and validate the transferability of the learned occupancy representation to real-world aerial environments.

\end{itemize}

\begin{figure*}[t]
\centering
\includegraphics[width=1\textwidth]{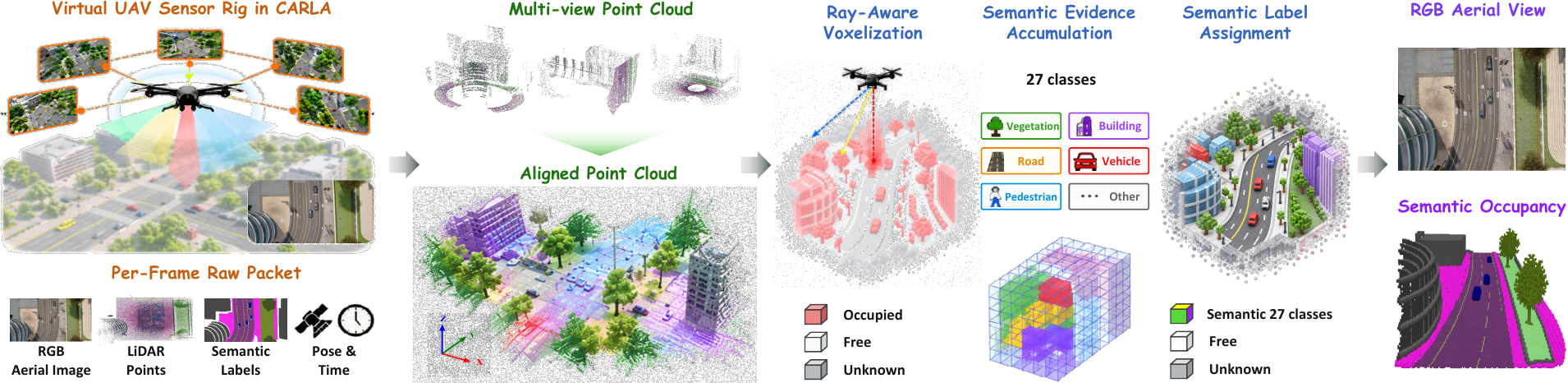} 
\caption{Overview of the AerialOcc-20K dataset generation pipeline. Multi-view UAV sensor observations are aligned and fused to generate dense semantic occupancy annotations with corresponding scene-level metadata.
}
\label{fig:AerialOcc-20K}
\end{figure*}

%% file: sec/2_related.tex
\section{Related Work}
\label{sec:related}

\subsection{3D Spatial Perception and Occupancy}

3D perception has evolved from object-centric localization toward scene-centric representation. 3D detection methods~\cite{lang2019pointpillars,yin2021center,liu2022petr,wang2022detr3d} localize object instances, while semantic segmentation methods such as PointNet~\cite{qi2017pointnet}, PointNet++~\cite{qi2017pointnet++}, and Cylinder3D~\cite{zhu2021cylindrical} assign labels to points or voxels. BEV-based methods~\cite{philion2020lift,li2023bevdepth,li2024bevformer,liang2022bevfusion} aggregate multi-view observations into a unified plane, but often compress vertical geometry.

In contrast, semantic scene completion~\cite{cao2022monoscene} infers voxelized geometry and semantics from partial observations, while visibility-aware occupancy prediction explicitly distinguishes occupied, free, and unobserved regions. Based on this scene-centric formulation, recent occupancy methods improve multi-view fusion~\cite{wei2023surroundocc,wang2024panoocc} and 3D spatial modeling~\cite{zhang2023occformer,ma2024cotr} for road scenes. However, existing methods largely assume stable ground-vehicle viewpoints, whereas UAV occupancy under varying altitude, orientation, and vertical scene structure remains underexplored.

\subsection{UAV Datasets and Benchmarks}

Early UAV benchmarks primarily focused on 2D visual perception, including object detection and tracking~\cite{du2018unmanned,liu2025boost,yao2024domain}, semantic segmentation~\cite{lyu2020uavid}, and cross-modal vehicle detection~\cite{sun2022drone}. Later datasets introduced geometric supervision: WildUAV~\cite{florea2021wilduav} provides depth and camera poses for depth estimation, while UAV3D~\cite{ye2024uav3d} introduces 3D bounding boxes for object-level detection and tracking. Despite increasingly diverse annotation formats, these datasets primarily provide depth, surface, or object-level supervision and lack voxel-level supervision that distinguishes occupied, free, and unobserved regions, limiting their use for scene-level occupancy learning and evaluation.

Recent aerial occupancy benchmarks have begun to address this gap. OccuFly~\cite{gross2026occufly} derives real-world aerial semantic occupancy annotations through 3D reconstruction for semantic scene completion, whereas SkyShield~\cite{gao2026skyshield} provides simulated occupancy supervision for low-altitude forward-view perception. Despite this progress, scene-centric occupancy remains less extensively benchmarked for UAVs than for ground vehicles, particularly across varied aerial environments.

%% file: sec/3_dataset.tex
\section{AerialOcc-20K Dataset}
\label{sec:dataset}

\subsection{Dataset Pipeline}
\label{sec:dataset-pipeline}
The overall pipeline of AerialOcc-20K is illustrated in Fig.~\ref{fig:AerialOcc-20K}, consisting of simulation data collection, reference-sensor alignment, and occupancy annotation generation.

\subsubsection{Simulation Data Collection.}

AerialOcc-20K comprises 20 simulated scenes across four CARLA towns with diverse spatial layouts. Town03, Town05, and Town10HD provide varied urban road networks and building structures, while Town07 introduces complementary rural environments. Additional scenes are collected in Town10HD to enhance the coverage of complex urban geometry and semantic structures.

\begin{figure}[t]
\centering
\includegraphics[width=\columnwidth]{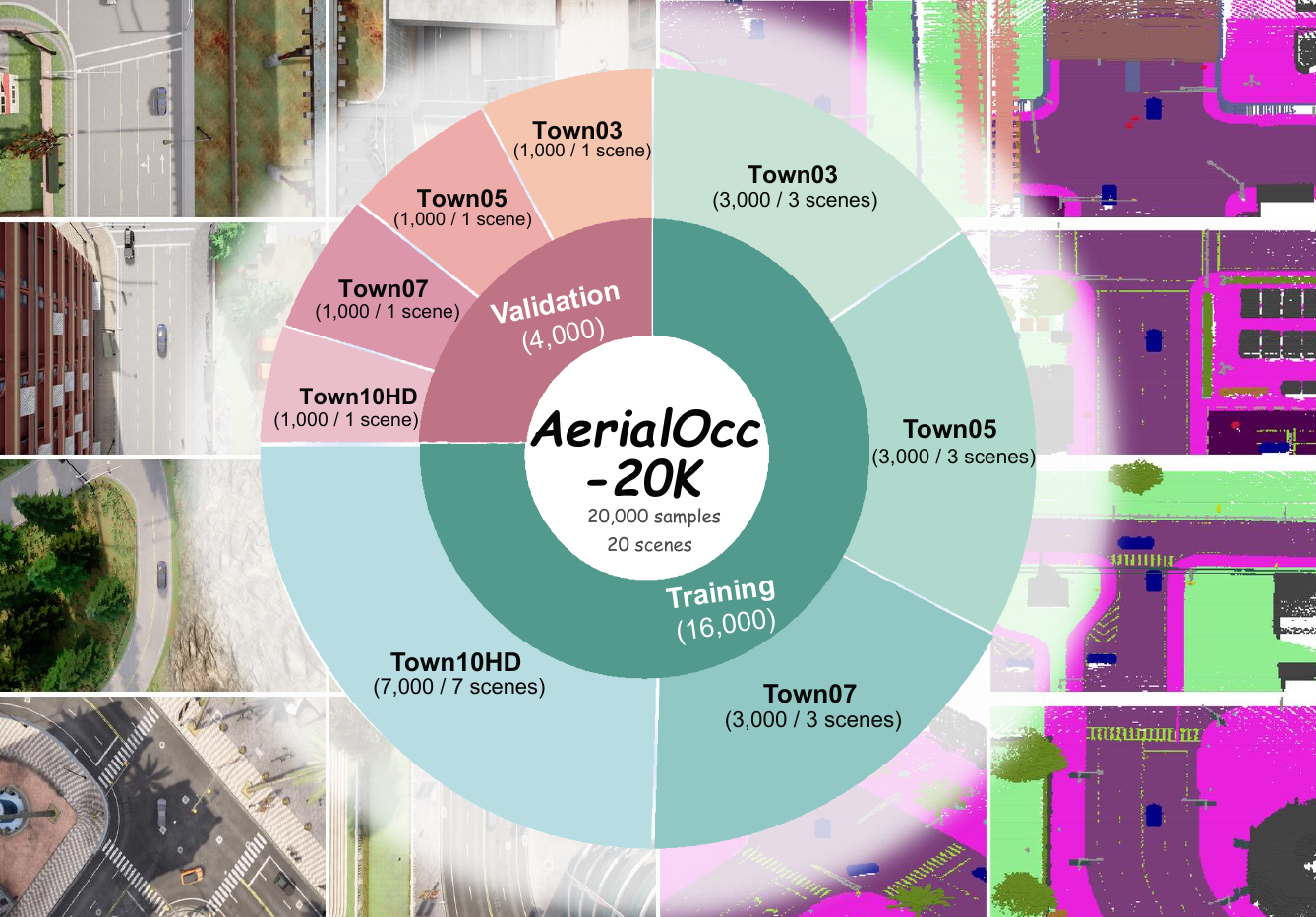}
\caption{Dataset statistics and composition of AerialOcc-20K. 
}
\label{fig:scene_composition}
\end{figure}

During data collection, CARLA operates in synchronous mode at \(10\,\mathrm{Hz}\), with RGB images and semantic LiDAR measurements matched using a unified frame identifier. We collect 1,000 frames per scene, yielding 20K samples in total. As shown in Fig.~\ref{fig:scene_composition}, the dataset is split into 16 training and 4 validation scenes, with a distinct collection trajectory for each scene. Each frame records an RGB image, 3D point coordinates, semantic and instance labels, dynamic object velocities, sensor poses, and timestamps.

\begin{table*}[t]
\centering
\caption{Comparison of terrestrial and aerial semantic occupancy
benchmarks. Classes exclude free and unobserved states.}
\label{tab:dataset_comparison}

\small
\renewcommand{\arraystretch}{0.95}
\setlength{\tabcolsep}{4pt}

\resizebox{\textwidth}{!}{%
\begin{tabular}{lccccccc}
\toprule
\multicolumn{1}{l}{\multirow{2}{*}{\textbf{Dataset}}}
& \multirow{2}{*}{\textbf{Publish}}
& \textbf{Camera}
& \textbf{\# of}
& \textbf{Depth}
& \textbf{Semantic}
& \textbf{Grid}
& \textbf{Voxel} \\

&
&
\textbf{Views}
&
\textbf{Samples}
&
\textbf{Maps}
&
\textbf{Classes}
&
\textbf{Dimensions}
&
\textbf{Size} \\

\midrule

\rowcolor{lightgray}
\multicolumn{8}{l}{\textit{Terrestrial Benchmarks}} \\

SemanticKITTI~\cite{behley2019semantickitti}
& ICCV 2019
& single
& 4,649
& $\times$
& 19
& $256\times256\times32$
& 0.20 \\

SSCBench-Waymo~\cite{li2024sscbench}
& IROS 2024
& multi
& 19,985
& $\times$
& 14
& $256\times256\times32$
& 0.20 \\

SSCBench-nuScenes~\cite{li2024sscbench}
& IROS 2024
& multi
& 34,078
& $\times$
& 16
& $256\times256\times32$
& 0.20 \\

SSCBench-KITTI-360~\cite{li2024sscbench}
& IROS 2024
& multi
& 12,865
& $\times$
& 19
& $256\times256\times32$
& 0.20 \\

OpenOccupancy~\cite{wang2023openoccupancy}
& ICCV 2023
& multi
& 34,149
& $\times$
& 16
& $512\times512\times40$
& 0.20 \\

Occ3D-nuScenes~\cite{tian2023occ3d}
& NeurIPS 2023
& multi
& 40,000
& $\times$
& 16
& $200\times200\times16$
& 0.40 \\

Occ3D-Waymo~\cite{tian2023occ3d}
& NeurIPS 2023
& multi
& 200,000
& $\times$
& 14
& $3200\times3200\times128$
& 0.05 \\

\midrule

\rowcolor{lightgray}
\multicolumn{8}{l}{\textit{Aerial Benchmarks}} \\

SkyShield~\cite{gao2026skyshield}
& arXiv 2026
& single
& 36,000
& $\times$
& 9
& $100\times100\times60$
& 0.50 \\

OccuFly~\cite{gross2026occufly}
& CVPR 2026
& single
& 20,611
& $\checkmark$
& 21
& $192\times128\times128$
& 0.50 \\

\textbf{AerialOcc-20K (Ours)}
& --
& single
& 20,000
& $\checkmark$
& 26
& $256\times256\times192$
& 0.20 \\

\bottomrule
\end{tabular}%
}

\end{table*}

\subsubsection{Reference-Sensor Alignment.}

To establish a common spatial basis, we use semantic LiDAR 0, as the reference frame and align all other LiDAR measurements to it. Let \(\mathbf{T}^{w}_{s,t}\) and \(\mathbf{T}^{w}_{0,t}\) denote the transformations from sensor \(s\) and LiDAR 0 to the CARLA world coordinate system at timestamp \(t\), respectively. The relative transformation from sensor \(s\) to the reference frame is then given by
\begin{equation}
    \mathbf{T}^{0}_{s,t}
    =
    \left(\mathbf{T}^{w}_{0,t}\right)^{-1}
    \mathbf{T}^{w}_{s,t}.
\end{equation}

Accordingly, each LiDAR point \(\mathbf{p}^{s}_{t}\) is mapped to the reference frame as
\begin{equation}
    \tilde{\mathbf{p}}^{0}_{s,t}
    =
    \mathbf{T}^{0}_{s,t}
    \tilde{\mathbf{p}}^{s}_{t},
\end{equation}
where \(\tilde{\mathbf{p}}^{s}_{t}\) and
\(\tilde{\mathbf{p}}^{0}_{s,t}\) denote the homogeneous coordinates of the point in the sensor and reference frames, respectively. Each point retains its semantic label during transformation, and the resulting frame-wise aligned point set provides a unified geometric basis for occupancy annotation.

\subsubsection{Occupancy Annotation Generation.}

Following alignment, we voxelize the multi-view semantic point clouds into a \(256\times256\times192\) Cartesian grid centered at the reference sensor. For each valid Lidar return, the endpoint retains its semantic label, while points sampled at \(1\,\mathrm{m}\) intervals along the preceding ray segment are labeled as free space. The resulting semantic and free-space evidence from all viewpoints is then accumulated in the shared grid. AerialOcc-20K defines 27 effective occupancy labels, comprising one observed free-space class and 26 non-empty semantic classes. 

Let \(A_t(v,c)\) denote the observation count of class \(c\) accumulated in voxel \(v\) at timestamp \(t\), where \(c=0\) represents observed free space and \(c\in\mathcal{C}_{\mathrm{sem}}\) represents a non-empty semantic class. Let \(S_t(v)=\sum_{c\in\mathcal{C}_{\mathrm{sem}}}A_t(v,c)\) denote the total non-empty semantic evidence in voxel \(v\). The voxel label \(y_t(v)\) is obtained by majority voting over the non-empty semantic classes:
\begin{equation}
y_t(v)=
\left\{
\begin{array}{ll}
\displaystyle
\mathop{\arg\max}_{c\in\mathcal{C}_{\mathrm{sem}}} A_t(v,c),
& S_t(v)>0, \\[4pt]
0,
& S_t(v)=0.
\end{array}
\right.
\end{equation}

Since \(y_t(v)=0\) conflates observed free space with unobserved space, we define a valid observation mask \(m_t(v)\), which is set to \(1\) if voxel \(v\) accumulates any semantic return or free-space sample and to \(0\) otherwise. Occupancy training and evaluation are restricted to voxels with \(m_t(v)=1\), excluding unobserved space from free-space supervision.

\subsection{Data Analysis}

As shown in Tab.~\ref{tab:dataset_comparison}, AerialOcc-20K combines a broad semantic vocabulary with the highest voxel count per sample among existing aerial semantic occupancy benchmarks. Beyond these, its scene-level annotations exhibit two complementary properties:

1) Semantic coverage. AerialOcc-20K defines 26 non-empty semantic classes spanning ground surfaces, built structures, natural elements, traffic infrastructure, and dynamic participants. Together, these categories form a unified voxel taxonomy, allowing the annotations to represent both scene structure and fine-grained local semantics.

2) Multi-altitude coverage. Each RGB frame is paired with a 3D occupancy annotation aggregated from 24 virtual semantic LiDAR sensors positioned at heights of \(3\)--\(32\,\mathrm{m}\) with multiple pitch angles. This multi-altitude rig integrates complementary viewpoints to improve coverage across observed regions and support geometrically aligned occupancy supervision.

%% file: sec/4_aerial.tex
\section{AerialOcc}
\label{sec:aerial}


\begin{figure*}[t]
\centering
\includegraphics[width=1\textwidth]{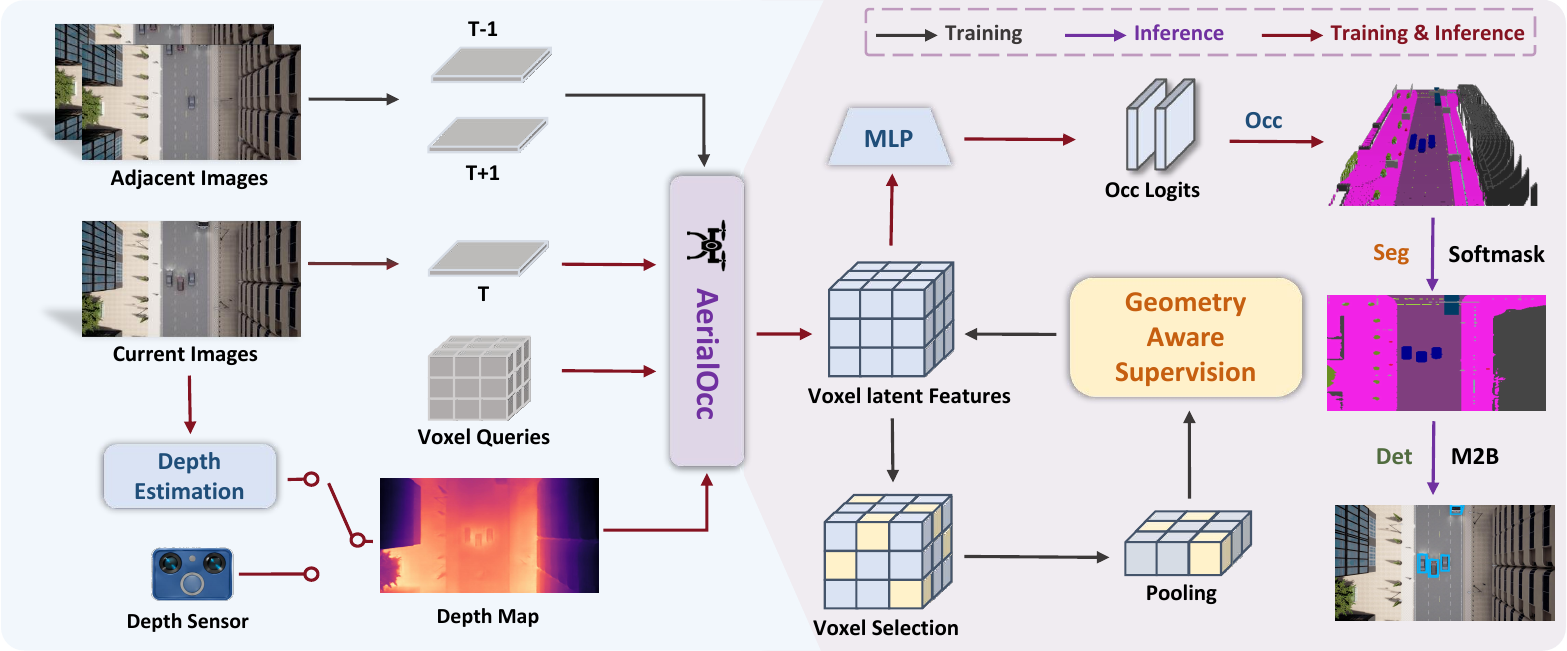} 
\caption{Overview of AerialOcc. Given adjacent and current aerial images, together with estimated or sensor-provided depth maps, AerialOcc constructs 3D voxel latent features through voxel queries and geometric feature encoding. The voxel features are processed by a MLP to predict semantic occupancy logits, while geometry aware supervision is applied to refine voxel feature. The refined occupancy volume serves as a unified 3D scene representation, from which semantic segmentation and object detection are derived through occupancy-based post-processing.}
\label{fig:Method}
\end{figure*}

\subsection{Problem Formulation}
\label{sec:problem_formulation}

We consider aerial perception from a temporal sequence of calibrated RGB observations centered at time \(t\). Let \(\mathcal{X}_t=(\mathcal{I}_t,\mathcal{K}_t,\mathcal{T}_t,D_t^{\mathrm{proj}})\) denote the input, where \(\mathcal{I}_t\) contains the adjacent RGB images, \(\mathcal{K}_t\) contains the camera intrinsics, \(\mathcal{T}_t\) contains the relative poses, and \(D_t^{\mathrm{proj}}\) denotes the projected LiDAR depth. Let \(\mathcal{Q}=\{\mathrm{ssc},\mathrm{seg},\mathrm{det}\}\) denote semantic scene completion, 3D semantic segmentation, and 3D object detection. Conventional perception systems predict each task using an independent task-specific function:
\begin{equation}
\hat{y}_t^{q}=F_{\theta_q}^{q}(\mathcal{X}_t),\quad q\in\mathcal{Q}.
\label{eq:task_specific_formulation}
\end{equation}

In contrast, AerialOcc treats semantic occupancy as the common representation for all perception tasks. It first predicts a semantic occupancy volume:
\begin{equation}
O_t=F_{\theta}(\mathcal{X}_t),\quad O_t\in\mathbb{R}^{K\times X\times Y\times Z},
\label{eq:aerialocc_formulation}
\end{equation}
where \(K\) includes one free-space class and \(K-1\) non-empty semantic classes, and \(X\), \(Y\), and \(Z\) denote the voxel-grid dimensions. All task outputs are then derived through task conversion functions:
\begin{equation}
\hat{y}_t^{q}=\Phi_q(O_t),\quad q\in\mathcal{Q}.
\label{eq:unified_output_formulation}
\end{equation}
Here, \(\Phi_{\mathrm{ssc}}\) returns the complete semantic occupancy volume, \(\Phi_{\mathrm{seg}}\) extracts voxel-wise semantic labels for occupied regions, and \(\Phi_{\mathrm{det}}\) separates class-specific voxel regions into object instances and converts them into oriented 3D bounding boxes. Thus, AerialOcc unifies multiple aerial perception tasks as different interpretations of the same semantic occupancy prediction.
\subsection{Temporal Aware Feature Fusion}

As illustrated in Fig.~\ref{fig:Method}, given three adjacent aerial observations \(\{I_{t-1},I_t,I_{t+1}\}\), AerialOcc extracts multi-scale image features using a shared encoder. By aggregating complementary visual information across the three observations, the model captures temporal context and spatial variations, providing richer visual representations for subsequent 3D geometric modeling.

Specifically, the shared encoder and FPN extract multi-scale features from the three observations, which are subsequently fed into a Temporal Feature Fusion module. For each observation, its feature is concatenated with the center-frame feature and the temporal mean feature, followed by convolutional fusion. The fused feature at the center timestamp is retained as the temporal representation:
\begin{equation}
F_t^{\mathrm{temp}}
=
\mathcal{F}_{\mathrm{temp}}
\left(
F_t\|F_t\|
\frac{1}{3}\sum_{i=t-1}^{t+1}F_i
\right),
\label{eq:temporal_fusion}
\end{equation}
where \(F_{t-1}\), \(F_t\), and \(F_{t+1}\) denote the extracted image features, \(\|\) denotes feature concatenation, and \(\mathcal{F}_{\mathrm{temp}}\) represents the convolution-based fusion operation. The resulting \(F_t^{\mathrm{temp}}\) integrates complementary information across the three observations and serves as the visual input for subsequent geometric modeling.

\subsection{Geometry Aware Occupancy Supervision}

Reconstructing reliable 3D geometry from aerial observations is challenging due to incomplete observations and depth ambiguity. To address this issue, AerialOcc incorporates LiDAR-projected depth as an additional geometric input. A depth estimation network predicts monocular depth distributions and image context features, while a depth encoder extracts geometric features from projected LiDAR depth. A depth aggregation module combines both sources to obtain the fused depth distribution \(P_t^{\mathrm{depth}}\).

We further employ a frozen VGGT model as a geometric teacher to provide geometric prior supervision during training. Its predicted depth maps are converted into Gaussian soft distributions to guide the student's depth distribution through KL-based distillation:
\begin{equation}
\mathcal{L}_{\mathrm{KD}}
=
D_{\mathrm{KL}}
\left(
P_t^{\mathrm{teacher}}
\parallel
P_t^{\mathrm{depth}}
\right),
\label{eq:depth_kd}
\end{equation}
where \(P_t^{\mathrm{teacher}}\) and \(P_t^{\mathrm{depth}}\) denote the teacher and student geometric depth distributions, respectively. VGGT is only used during training and is not required for inference.

The temporal features and geometric cues are subsequently integrated into a unified voxel representation. Specifically, LSS generates coarse 3D priors, while projected LiDAR measurements provide voxel proposals. Learnable voxel queries \(Q_v\) aggregate visual and geometric information through depth-guided attention, followed by local-global feature aggregation. The resulting features are mapped to voxel-wise semantic occupancy logits:
\begin{equation}
\begin{aligned}
V_t &= \mathcal{E}_{\mathrm{occ}}
\left(Q_v,F_t^{\mathrm{temp}},P_t^{\mathrm{depth}},D_t^{\mathrm{proj}}\right),\\
O_t &= H_{\mathrm{occ}}(V_t),
\end{aligned}
\label{eq:voxel_feature}
\end{equation}
where \(D_t^{\mathrm{proj}}\) denotes the projected LiDAR depth, \(V_t\in\mathbb{R}^{C\times X\times Y\times Z}\) represents the voxel features, and \(O_t\in\mathbb{R}^{K\times X\times Y\times Z}\) denotes the occupancy logits, with \(K=27\). Let \(P_t=\operatorname{Softmax}(O_t)\) and \(Y_t\) denote the predicted probabilities and voxel annotations, respectively. \(P_t^{\mathrm{occ}}\) and \(Y_t^{\mathrm{occ}}\) represent their binary occupied/free forms. The occupancy objective is
\begin{equation}
\mathcal{L}_{\mathrm{occ}}
=
\mathcal{L}_{\mathrm{CE}}(O_t,Y_t)
+
\mathcal{L}_{\mathrm{sem}}(P_t,Y_t)
+
\mathcal{L}_{\mathrm{geo}}(P_t^{\mathrm{occ}},Y_t^{\mathrm{occ}}),
\label{eq:occ_loss}
\end{equation}
where the three terms optimize voxel-wise classification, class-wise semantic accuracy, and occupied/free geometric consistency, respectively. The overall objective combines occupancy supervision and geometric distillation:
\begin{equation}
\mathcal{L}_{\mathrm{total}}
=
\mathcal{L}_{\mathrm{occ}}
+
\lambda_{\mathrm{KD}}\mathcal{L}_{\mathrm{KD}},
\label{eq:total_loss}
\end{equation}
where \(\lambda_{\mathrm{KD}}\) controls the contribution of geometric prior distillation.















\begin{table}[t]
\centering
\caption{Semantic scene completion results on the AerialOcc-20K validation set.}
\label{tab:carla_occ_results}

\normalsize
\renewcommand{\arraystretch}{0.85}

\def\OccTable{%
\begin{tabular}{l|c|cc}
\toprule
\textbf{Method}
& \textbf{Input}
& \textbf{IoU}$\uparrow$
& \textbf{mIoU}$\uparrow$ \\
\midrule

CGFormer~\cite{yu2024context}
& C+L
& \underline{60.15}
& 3.52 \\

M3Net~\cite{chen2025m3net}
& C+L
& 44.51
& 8.07 \\

PanoOcc~\cite{wang2024panoocc}
& C
& 56.29
& \underline{19.89} \\

\rowcolor{lightblue}
AerialOcc (Ours)
& C+L
& \textbf{69.02}
& \textbf{29.93} \\

\bottomrule
\end{tabular}%
}

\setlength{\tabcolsep}{0pt}
\sbox0{\OccTable}
\setlength{\tabcolsep}{\dimexpr(\columnwidth-\wd0)/8\relax}

\OccTable

\end{table}

\begin{table}[t]
\centering
\caption{3D semantic segmentation results on the AerialOcc-20K validation set.}
\label{tab:voxel_seg}

\normalsize
\renewcommand{\arraystretch}{0.85}

\def\SegTable{%
\begin{tabular}{l|c|cccc}
\toprule
\textbf{Method}
& \textbf{Input}
& \textbf{mIoU}$\uparrow$
& \textbf{OA}$\uparrow$
& \textbf{oIoU}$\uparrow$
& \textbf{F1}$\uparrow$ \\
\midrule

PointNet~\cite{qi2017pointnet}
& L
& 2.22
& 6.86
& 6.70
& 3.88 \\

PointNet++~\cite{qi2017pointnet++}
& L
& 1.51
& 5.90
& 5.70
& 2.69 \\

Cylinder3D~\cite{zhu2021cylindrical}
& L
& 2.89
& 7.83
& 7.71
& 5.19 \\

SparseUNet~\cite{2016SparseUNet}
& L
& 3.72
& 8.70
& 8.65
& 6.52 \\

BEV-Seg3D-Net~\cite{Zou2021EfficientUP}
& C+L
& 22.66
& \textbf{75.75}
& 60.96
& 30.09 \\

\midrule

M3Net~\cite{chen2025m3net}
& C+L
& 12.04
& 51.55
& 39.56
& 18.21 \\

PanoOcc~\cite{wang2024panoocc}
& C
& \underline{25.51}
& 71.40
& \underline{66.60}
& \underline{33.37} \\

\rowcolor{lightblue}
AerialOcc (Ours)
& C+L
& \textbf{36.79}
& \underline{75.70}
& \textbf{71.62}
& \textbf{44.27} \\

\bottomrule
\end{tabular}%
}

\setlength{\tabcolsep}{0pt}
\sbox0{\SegTable}
\setlength{\tabcolsep}{\dimexpr(\columnwidth-\wd0)/12\relax}

\SegTable

\end{table}

\subsection{Unified Perception from Occupancy}

During inference, AerialOcc uses the predicted occupancy logits \(O_t\in\mathbb{R}^{K\times X\times Y\times Z}\) as a unified representation for multiple perception tasks. For each voxel \(v\) and semantic class \(c\in\mathcal{C}=\{1,\ldots,K\}\), the class probability is computed as
\begin{equation}
P_t(c\mid v)=
\frac{\exp\left(O_t(c,v)\right)}
{\sum_{k=1}^{K}\exp\left(O_t(k,v)\right)},
\label{eq:occupancy_probability}
\end{equation}
where \(\mathcal{C}\) denotes the semantic occupancy class set. Semantic scene completion directly uses the predicted occupancy volume, while 3D semantic segmentation assigns a semantic label to each voxel:
\begin{equation}
\hat{S}_t(v)=\mathop{\arg\max}_{c\in\mathcal{C}}P_t(c\mid v),
\label{eq:occupancy_segmentation}
\end{equation}
where \(\hat{S}_t(v)\) denotes the predicted voxel-wise semantic label.

For 3D object detection, let \(\mathcal{C}_{\mathrm{obj}}\subseteq\mathcal{C}\) denote the detectable object categories. Class-specific voxel sets are extracted from the predicted semantic labels:
\begin{equation}
\mathcal{V}_t^c=
\left\{v\mid\hat{S}_t(v)=c\right\},
\quad c\in\mathcal{C}_{\mathrm{obj}},
\label{eq:class_voxel_mask}
\end{equation}
where \(\mathcal{V}_t^c\) denotes the voxel set belonging to category \(c\). We perform class-wise 3D connected-component (3D-CC) analysis with 26-neighborhood connectivity and discard components containing fewer than eight voxels. The remaining instance-wise voxel components \(\mathcal{R}_t\) are converted into oriented 3D bounding boxes through Mask-to-Bbox (M2B)~\cite{liu2024remoteclip,yao2025remotesam}. Specifically, the XY extent is determined by a minimum-area rotated rectangle fitted to projected voxel-cell corners, while the Z extent is obtained from voxel boundaries. Each box is assigned a confidence score based on the mean class probability of its highest-confidence 20\% voxels. The conversion process is summarized as
\begin{equation}
\{\mathcal{V}_t^c\}_{c\in\mathcal{C}_{\mathrm{obj}}}
\xrightarrow{\mathrm{3D\text{-}CC}}
\mathcal{R}_t
\xrightarrow{\mathrm{M2B}}
\hat{B}_t,
\label{eq:occupancy_detection}
\end{equation}
where \(\mathcal{R}_t\) denotes the retained instance-wise voxel components and \(\hat{B}_t\) represents the resulting oriented 3D bounding boxes. Thus, all three perception tasks are derived from the same occupancy prediction without additional learnable task-specific heads.

%% file: sec/5_experiments.tex
\begin{table*}[t]
\centering
\caption{3D object detection results on AerialOcc-20K for the car class and all 26 classes.}
\label{tab:detection}

\normalsize
\renewcommand{\arraystretch}{0.85}

\def\DetectionTable{%
\begin{tabular}{l|c|cccc|cccc}
\toprule
\multirow{2}{*}{\textbf{Method}}
& \multirow{2}{*}{\textbf{Input}}
& \multicolumn{4}{c|}{\textbf{Car}}
& \multicolumn{4}{c}{\textbf{26 Classes}} \\

\cline{3-6}
\cline{7-10}

&
& \rule{0pt}{2ex}\textbf{AP@0.25}$\uparrow$
& \textbf{AP@0.50}$\uparrow$
& \textbf{mATE}$\downarrow$
& \textbf{mASE}$\downarrow$
& \textbf{mAP@0.25}$\uparrow$
& \textbf{mAP@0.50}$\uparrow$
& \textbf{mATE}$\downarrow$
& \textbf{mASE}$\downarrow$ \\
\midrule

PETR~\cite{liu2022petr}
& C
& 28.30
& 27.17
& \underline{0.1401}
& 0.1923
& 1.93
& 1.60
& 0.9543
& 0.8284 \\

BEVFusion~\cite{liang2022bevfusion}
& C+L
& 21.30
& 5.60
& 0.2942
& 0.4527
& 0.83
& 0.22
& 0.9729
& 0.9026 \\

DETR3D~\cite{wang2022detr3d}
& C
& 32.24
& 30.12
& 0.2417
& \textbf{0.1312}
& 2.94
& \underline{2.17}
& 0.9254
& \textbf{0.7095} \\

M3Net~\cite{chen2025m3net}
& C+L
& 6.43
& 2.81
& 0.7780
& 0.6224
& 0.37
& 0.11
& 0.9746
& 0.8680 \\

PanoOcc~\cite{wang2024panoocc}
& C
& \textbf{35.36}
& \textbf{30.73}
& 0.4629
& 0.2185
& \underline{3.64}
& 1.95
& \underline{0.9191}
& 0.7208 \\

\rowcolor{lightblue}
AerialOcc (Ours)
& C+L
& \underline{34.98}
& \underline{30.56}
& \textbf{0.0378}
& \underline{0.1592}
& \textbf{6.49}
& \textbf{4.03}
& \textbf{0.7702}
& \underline{0.7117} \\

\bottomrule
\end{tabular}%
}

\setlength{\tabcolsep}{0pt}
\sbox0{\DetectionTable}
\setlength{\tabcolsep}{\dimexpr(\textwidth-\wd0)/20\relax}

\DetectionTable

\end{table*}

\section{Experiments}
\label{sec:experiments}
\subsection{Dataset and Metrics}
\subsubsection{Datasets}
We evaluate AerialOcc on AerialOcc-20K, UAV3D, and OccuFly. AerialOcc-20K is used for model training and evaluation, UAV3D measures transferability to real aerial viewpoints, and OccuFly evaluates sim-to-real generalization under different target-domain supervision.

\subsubsection{Metrics}
For semantic scene completion, we report occupancy IoU and semantic mIoU. 3D semantic segmentation is evaluated using mIoU, overall IoU (oIoU), overall accuracy (OA), and F1 score. For 3D object detection, we report AP@0.25 and AP@0.50 for the car class, and mAP@0.25 and mAP@0.50 across all 26 classes, together with mean translation error (mATE) and mean scale error (mASE).

\subsection{Implementation Details}

We adopt EfficientNet-B7 with an FPN as the shared image encoder and use Temporal Feature Fusion to aggregate features from three adjacent observations. AerialOcc is optimized using AdamW for 25,000 iterations, with an initial learning rate of \(3\times10^{-4}\) and a cosine learning rate schedule. The batch size is set to 1. All experiments are conducted on two NVIDIA GeForce RTX 4090 GPUs with 24 GB.


\begin{table}[t]
\centering
\caption{
Semantic scene completion evaluation on the OccuFly validation set.
}
\label{tab:occufly_transfer}
\setlength{\tabcolsep}{8pt}
\renewcommand{\arraystretch}{0.85}

\begin{tabular}{c|l|cc}
\toprule

\textbf{Setting}
&
\textbf{Method}
&
\textbf{IoU}$\uparrow$
&
\textbf{mIoU}$\uparrow$
\\

\midrule

\multirow{2}{*}{\centering Real}
&
CGFormer~\cite{yu2024context}
&
24.66
&
0.80
\\

&
AerialOcc (Ours)
&
\textbf{44.49}
&
\underline{5.49}
\\

\midrule

\multirow{3}{*}{\centering Sim-to-real}

&
Zero-shot (Ours)
&
15.54
&
0.02
\\

&
10\% FT (Ours)
&
33.39
&
5.44
\\

&
100\% FT (Ours)
&
\underline{41.18}
&
\textbf{5.64}
\\

\bottomrule

\end{tabular}

\end{table}


\begin{figure}[t]
\centering
\includegraphics[width=\columnwidth]{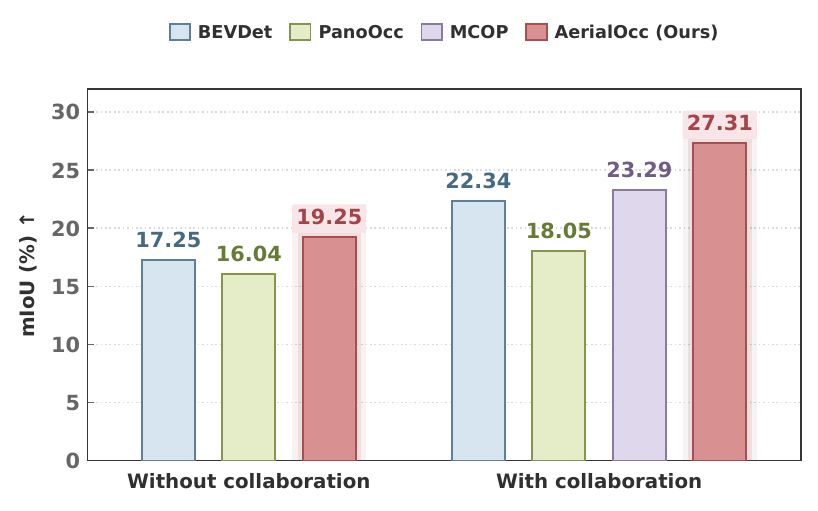}
\caption{Occupancy prediction results on UAV3D with and without collaboration.}
\label{fig:uav3d-collaboration}
\end{figure}

\begin{table}[t]
\centering
\caption{Ablation study on AerialOcc-20K validation set. G.S. and T.F. denote geometry supervision and temporal fusion.}
\label{tab:ablation_geometry}
\renewcommand{\arraystretch}{1.1}
\setlength{\tabcolsep}{10pt}
\begin{tabular}{ccc|cc}
\toprule
\textbf{CGFormer}
&
\textbf{G.S.}
&
\textbf{T.F.}
&
\textbf{IoU}$\uparrow$
&
\textbf{mIoU}$\uparrow$
\\
\midrule
$\checkmark$
&
&
&
60.15
&
3.52
\\
$\checkmark$
&
$\checkmark$
&
&
68.51
&
26.62
\\
$\checkmark$
&
&
$\checkmark$
&
68.59
&
26.91
\\
\rowcolor{lightblue}
$\checkmark$
&
$\checkmark$
&
$\checkmark$
&
\textbf{69.02}
&
\textbf{29.93}
\\
\bottomrule
\end{tabular}

\end{table}

\begin{table}[t]
\centering
\caption{Sensitivity analysis of the geometric distillation weight
\(\lambda_{\mathrm{KD}}\) on the AerialOcc-20K validation set.}
\label{tab:lambda_kd}

\renewcommand{\arraystretch}{0.85}

\def\KDTable{%
\begin{tabular}{c|ccc}
\toprule
\(\lambda_{\mathrm{KD}}\)
&
\textbf{OCC mIoU}$\uparrow$
&
\textbf{SEG mIoU}$\uparrow$
&
\textbf{DET mAP@0.25}$\uparrow$
\\
\midrule

0.00
& 26.91
& 27.48
& 4.65
\\

0.25
& 29.09
& 36.51
& 6.37
\\

0.50
& \underline{29.11}
& \underline{36.53}
& \underline{6.48}
\\

1.00
& \textbf{29.93}
& \textbf{36.79}
& \textbf{6.49}
\\

2.00
& 29.08
& 36.50
& 6.46
\\

\bottomrule
\end{tabular}%
}

\setlength{\tabcolsep}{0pt}
\sbox0{\KDTable}
\setlength{\tabcolsep}{\dimexpr(\columnwidth-\wd0)/8\relax}

\KDTable

\end{table}

\subsection{Main Results}
To accommodate the input requirements of different methods, we adapt the input data accordingly while preserving their respective model architectures. C denotes RGB camera images, L denotes LiDAR data, and C+L denotes the combination of camera and LiDAR inputs.

\begin{figure*}[t]
\centering
\includegraphics[width=1\textwidth]{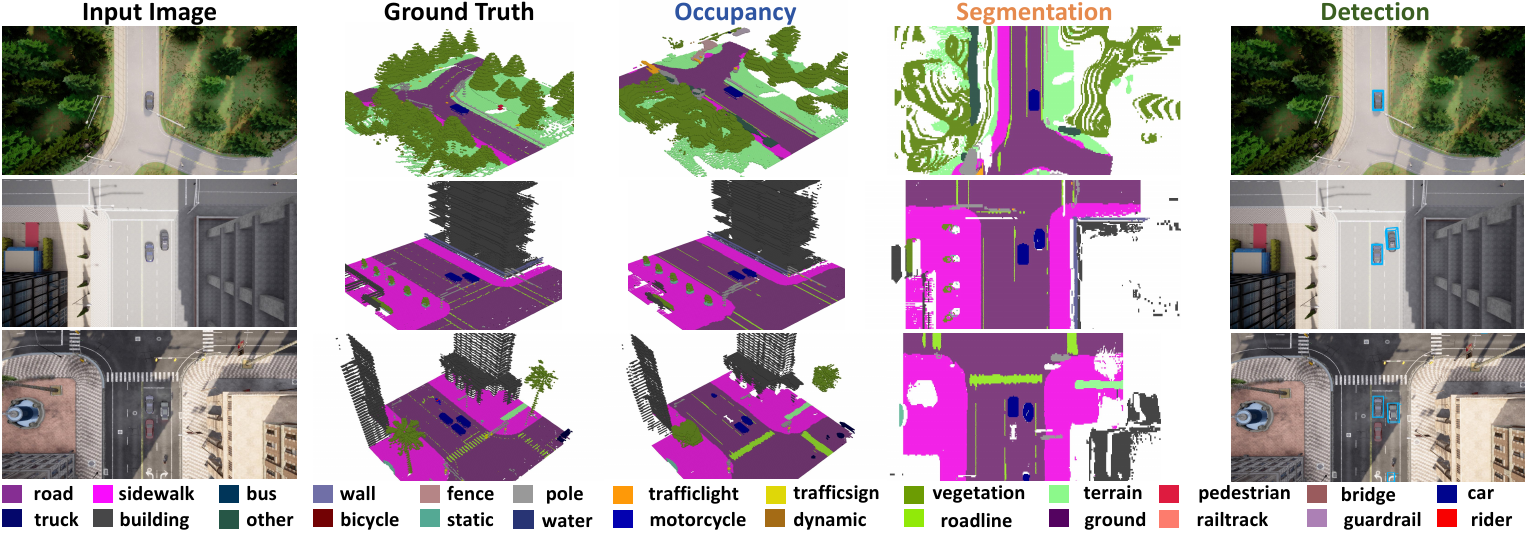}
\caption{Inference examples of AerialOcc on 3 visual-centric tasks.}
\label{fig:visualization}
\end{figure*}

\subsubsection{Performance on Semantic Scene Completion}
We evaluate AerialOcc on AerialOcc-20K and UAV3D. As reported in Tab.~\ref{tab:carla_occ_results}, AerialOcc achieves 69.02\% IoU and 29.93\% mIoU, outperforming CGFormer by 8.87 and 26.41 percentage points, respectively. As shown in Fig.~\ref{fig:uav3d-collaboration}, BEVDet~\cite{huang2021bevdet} is adapted for occupancy prediction using FlashOcc~\cite{yu2023flashocc}, while Where2comm~\cite{hu2022where2comm} is introduced to enable collaborative perception. AerialOcc achieves 19.25\% mIoU without collaboration and further improves to 27.31\% with collaboration, surpassing MCOP. These results demonstrate the effectiveness of AerialOcc in aerial semantic scene completion and collaborative occupancy prediction.

\subsubsection{Performance on 3D Semantic Segmentation}

We further evaluate AerialOcc against point-based and occupancy-based segmentation methods. As shown in Tab.~\ref{tab:voxel_seg}, AerialOcc achieves an mIoU of 36.79\% and an F1 score of 44.27\%. Although BEV-Seg3D-Net achieves a higher OA, AerialOcc delivers competitive overall segmentation performance without introducing an additional segmentation head. These results demonstrate that the unified occupancy representation can effectively preserve fine-grained semantic information and support dense 3D segmentation alongside scene completion.

\subsubsection{Performance on 3D Object Detection}

To further validate the detection performance of AerialOcc, we compare it with representative detection and occupancy-based methods. As shown in Tab.~\ref{tab:detection}, AerialOcc achieves the best mAP across all 26 classes, while its car detection performance is close to that of methods equipped with dedicated detection heads. This difference may be partly attributed to the limited representation of small and densely distributed instances in voxel-based occupancy predictions, despite their strong overall scene representation. Nevertheless, these results demonstrate that a unified occupancy representation can support competitive multi-class 3D detection without an additional learnable detection head.

\subsection{Further Analysis}
\subsubsection{Sim-to-Real}
We evaluate the sim-to-real transferability of AerialOcc on the real-world OccuFly dataset, with results reported in Tab.~\ref{tab:occufly_transfer}. Under the real-data setting, AerialOcc achieves 44.49\% IoU and 5.49\% mIoU. Starting from the AerialOcc-20K pretrained model, we randomly sample 10\% of the target-domain data from each flight altitude for fine-tuning, achieving 5.44\% mIoU. This result approaches the 5.64\% obtained with full target-domain fine-tuning, demonstrating the transferability of AerialOcc-20K pretraining under limited target-domain supervision.

\subsubsection{Qualitative Analysis}
We present qualitative results of AerialOcc in Fig.~\ref{fig:visualization}, including successful predictions and failure cases. Overall, the unified occupancy representation supports both 3D semantic segmentation and object detection. However, both tasks depend on the quality of occupancy predictions. As shown in the third row, missing car regions lead to segmentation errors and missed detections. 

\subsection{Ablation Studies}

\subsubsection{Temporal Fusion and Geometry Supervision}
We evaluate temporal fusion and geometry supervision in Tab.~\ref{tab:ablation_geometry}. Both components independently yield substantial improvements, while their combination achieves the best performance of 69.02\% IoU and 29.93\% mIoU. This confirms that explicit geometry and temporal observations provide complementary cues for occupancy learning.

\subsubsection{Sensitivity to Geometric Distillation}

We evaluate the sensitivity to the geometric distillation weight \(\lambda_{\mathrm{KD}}\) in Tab.~\ref{tab:lambda_kd}. The results remain relatively stable across non-zero weight settings, indicating that AerialOcc is robust to moderate variations in geometric distillation strength.

%% file: sec/6_conclusion.tex
\section{Conclusion}
\label{sec:conclusion}


In this work, we present AerialOcc, a unified aerial 3D perception model that addresses the reliance of existing methods on task-specific prediction heads through an occupancy-centric paradigm. To support this paradigm, we construct AerialOcc-20K, a large-scale semantic occupancy dataset comprising 20K aerial samples with dense voxel-wise annotations over 27 occupancy classes and diverse altitudes and viewpoints. Extensive evaluations demonstrate that AerialOcc effectively supports semantic scene completion, 3D semantic segmentation, and 3D object detection from a single occupancy prediction, while exhibiting transferability to real aerial datasets. Our work demonstrates that occupancy-centric architectures can serve as a flexible and scalable interface for unified aerial 3D perception.


%% file: sec/X_suppl.tex
\section*{A. More Dataset Composition Details}

AerialOcc-20K comprises 20 scenes generated across four CARLA maps: Town03, Town05, Town07, and Town10HD. Table~\ref{tab:supp_scene_composition} presents their allocation across the training and validation splits. The benchmark contains 16,000 training samples and 4,000 validation samples. All 20,000 samples pass the synchronization integrity check, with complete RGB observations and measurements from all 24 semantic LiDAR sensors, and no missing sensor frames or out-of-range semantic labels.

As shown in Table~\ref{tab:sensor_rig}, we configure a virtual aerial sensor rig referenced to the CARLA ego actor for data collection. The rig comprises 24 semantic LiDAR sensors distributed across relative heights of \(3\)--\(32\,\mathrm{m}\) and one downward-facing RGB camera at \(35\,\mathrm{m}\). Each LiDAR uses 64 channels with a maximum range of \(50\,\mathrm{m}\), while the RGB camera synchronously captures images at a resolution of \(1920\times1080\).
\begin{table}[h]
\centering
\begin{minipage}[t]{0.53\columnwidth}
\centering
\caption{Dataset statistics and composition of AerialOcc-20K.}
\label{tab:supp_scene_composition}
\vspace{2pt}
\setlength{\tabcolsep}{2.5pt}
\renewcommand{\arraystretch}{1.05}

\begin{tabular}{l|ccc}
\toprule
\textbf{Town}
& \textbf{Scene IDs}
& \textbf{\# Scenes}
& \textbf{\# Samples} \\
\midrule

\rowcolor{lightgray}
\multicolumn{4}{l}{\textit{Training}} \\

Town03 & 01, 04, 07 & 3 & 3,000 \\
Town05 & 00, 03, 06 & 3 & 3,000 \\
Town07 & 02, 05, 08 & 3 & 3,000 \\
Town10HD & 09--15 & 7 & 7,000 \\

\midrule
\rowcolor{lightgray}
\multicolumn{4}{l}{\textit{Validation}} \\

Town03 & 17 & 1 & 1,000 \\
Town05 & 16 & 1 & 1,000 \\
Town07 & 18 & 1 & 1,000 \\
Town10HD & 19 & 1 & 1,000 \\

\midrule
\textbf{Total}
& 00--19
& \textbf{20}
& \textbf{20,000} \\

\bottomrule
\end{tabular}
\end{minipage}
\hfill
\begin{minipage}[t]{0.43\columnwidth}
\centering
\caption{Configuration of the virtual aerial sensor rig. All groups except RGB denote semantic LiDAR sensors. Heights are relative to the CARLA ego actor.}
\label{tab:sensor_rig}
\vspace{2pt}
\setlength{\tabcolsep}{3pt}
\renewcommand{\arraystretch}{1.05}

\begin{tabular}{l|cccc}
\toprule
\textbf{Group}
& \textbf{Num.}
& \textbf{H (m)}
& \textbf{Pitch}
& \textbf{View} \\
\midrule
RGB & 1 & 35 & \(-90^\circ\) & Nadir \\
Ref. & 1 & 30 & \(0^\circ\) & Horiz. \\
High & 5 & 28--32 & \(-90^\circ\) & Nadir \\
Mid & 4 & 16--22 & \(-45^\circ\) & Oblique \\
Low & 6 & 8--12 & \(0^\circ\) & Horiz. \\
Near-16 & 2 & 16 & \(-90^\circ\) & Nadir \\
Near-12 & 2 & 12 & \(-90^\circ\) & Nadir \\
Ped-F & 2 & 3.6 & \(-10^\circ\) & Front \\
Ped-L & 2 & 3 & \(-8^\circ\) & Lateral \\
\bottomrule
\end{tabular}
\end{minipage}
\end{table}

\section*{B. More Method Details}

This section provides additional details of the training and inference procedures of AerialOcc. During training, AerialOcc learns a unified semantic occupancy representation through voxel-wise supervision and geometry-guided depth distillation. During inference, semantic scene completion, 3D semantic segmentation, and 3D object detection are derived from the same occupancy prediction through task-specific readout and geometric conversion.

\subsection*{B.1 Training Procedure}

Given three adjacent RGB observations, AerialOcc first extracts multi-scale image features using a shared encoder and FPN. The Temporal Feature Fusion module aggregates these features to obtain the temporal representation \(F_t^{\mathrm{temp}}\). Meanwhile, the depth estimation network predicts monocular depth distributions and image context features, which are combined with geometric features extracted from projected LiDAR depth to obtain the fused depth distribution \(P_t^{\mathrm{depth}}\).

During training, a frozen VGGT model provides geometric depth priors. Its predicted depth maps are converted into Gaussian soft distributions to supervise the fused depth distribution through KL divergence. The temporal features and geometric cues are subsequently aggregated by learnable voxel queries to construct the unified voxel representation \(V_t\). An occupancy prediction head maps these features to voxel-wise semantic occupancy logits \(O_t\).

The original semantic occupancy annotations are generated on a \(256\times256\times192\) voxel grid. To match the model output resolution, we perform 3D voxel resampling based on physical coordinates. Specifically, each source voxel is mapped to the target grid according to its physical center. Source voxels outside the target range are discarded, while their semantic labels are aggregated to construct the target annotations. The corresponding validity masks are aggregated using an OR operation, such that a target voxel is considered valid if at least one mapped source voxel contains valid observations.

The training objective combines occupancy supervision and geometric knowledge distillation:
\begin{equation}
\mathcal{L}_{\mathrm{total}}
=
\mathcal{L}_{\mathrm{occ}}
+
\lambda_{\mathrm{KD}}\mathcal{L}_{\mathrm{KD}},
\label{eq:supp_total_loss}
\end{equation}
where \(\mathcal{L}_{\mathrm{occ}}\) consists of voxel-wise classification, semantic, and geometric consistency losses, while \(\mathcal{L}_{\mathrm{KD}}\) transfers geometric priors from the frozen VGGT teacher. The validity mask is used to exclude unobserved voxels from occupancy supervision. The complete training procedure is summarized in Algorithm~\ref{alg:supp_training}.

\begin{algorithm}[t]
\small
\caption{Training Procedure of AerialOcc}
\label{alg:supp_training}
\begin{algorithmic}[1]
\REQUIRE Three adjacent RGB images, camera intrinsics
\(\mathcal{K}_t\), relative poses \(\mathcal{T}_t\),
projected LiDAR depth \(D_t^{\mathrm{proj}}\),
raw annotations \(Y_t^{\mathrm{raw}}\), observation counts
\(N_t^{\mathrm{raw}}\), and frozen VGGT teacher

\FOR{each training iteration}
    \STATE Resample source voxel annotations to the target grid
    using physical-coordinate mapping; aggregate validity masks using OR
    \STATE Extract multi-scale image features and fuse temporal
    information to obtain \(F_t^{\mathrm{temp}}\)
    \STATE Estimate monocular depth and encode projected LiDAR depth
    \STATE Aggregate geometric features to obtain \(P_t^{\mathrm{depth}}\)
    \STATE Generate VGGT Gaussian depth targets and compute
    \(\mathcal{L}_{\mathrm{KD}}\)
    \STATE Construct voxel features
    \(V_t=\mathcal{E}_{\mathrm{occ}}(Q_v,F_t^{\mathrm{temp}},
    P_t^{\mathrm{depth}},D_t^{\mathrm{proj}})\)
    \STATE Predict occupancy logits
    \(O_t=H_{\mathrm{occ}}(V_t)\) and probabilities
    \(P_t=\operatorname{Softmax}(O_t)\)
    \STATE Compute \(\mathcal{L}_{\mathrm{occ}}\) using valid voxels
    and the corresponding occupancy annotations
    \STATE Optimize
    \(\mathcal{L}_{\mathrm{total}}
    =\mathcal{L}_{\mathrm{occ}}
    +\lambda_{\mathrm{KD}}\mathcal{L}_{\mathrm{KD}}\)
    using AdamW
\ENDFOR
\end{algorithmic}
\end{algorithm}

\begin{algorithm}[t]
\small
\caption{Unified Inference Procedure of AerialOcc}
\label{alg:supp_inference}
\begin{algorithmic}[1]
\REQUIRE Three adjacent RGB images, camera intrinsics
\(\mathcal{K}_t\), relative poses \(\mathcal{T}_t\),
and projected LiDAR depth \(D_t^{\mathrm{proj}}\)
\ENSURE Semantic occupancy, voxel-wise semantic labels,
and oriented 3D bounding boxes

\STATE Extract and fuse multi-scale temporal image features
to obtain \(F_t^{\mathrm{temp}}\)
\STATE Estimate monocular depth and aggregate it with
projected LiDAR features to obtain \(P_t^{\mathrm{depth}}\)
\STATE Construct voxel features
\(V_t=\mathcal{E}_{\mathrm{occ}}(Q_v,F_t^{\mathrm{temp}},
P_t^{\mathrm{depth}},D_t^{\mathrm{proj}})\)
\STATE Predict occupancy logits
\(O_t=H_{\mathrm{occ}}(V_t)\) and probabilities
\(P_t=\operatorname{Softmax}(O_t)\)
\STATE Obtain semantic occupancy and voxel labels
\(\hat{S}_t(v)=\arg\max_{c\in\mathcal{C}}P_t(c\mid v)\)
\STATE Initialize detection set \(\hat{B}_t=\emptyset\)

\FOR{each detectable category \(c\in\mathcal{C}_{\mathrm{obj}}\)}
    \STATE Extract \(\mathcal{V}_t^c=\{v\mid\hat{S}_t(v)=c\}\)
    \STATE Apply 26-neighborhood 3D-CC and discard
    components with fewer than eight voxels
    \FOR{each retained component \(R\)}
        \STATE Apply M2B to obtain an oriented 3D box;
        calculate confidence from its highest-confidence 20\% voxels
        \STATE Add the box to \(\hat{B}_t\)
    \ENDFOR
\ENDFOR
\STATE \textbf{return} semantic occupancy, \(\hat{S}_t(v)\),
and \(\hat{B}_t\)
\end{algorithmic}
\end{algorithm}

\subsection*{B.2 Unified Inference Procedure}

During inference, AerialOcc predicts semantic occupancy logits from the input RGB observations and projected LiDAR depth. The logits are converted into class probabilities using softmax, and voxel-wise semantic labels are obtained through the argmax operation. The predicted occupancy volume directly provides semantic scene completion and 3D semantic segmentation.

For 3D object detection, class-specific voxel sets are extracted from the predicted semantic labels. Class-wise 3D connected-component (3D-CC) analysis with 26-neighborhood connectivity is applied to separate individual instances, and components containing fewer than eight voxels are discarded. Each remaining component is converted into an oriented 3D bounding box using geometric Mask-to-Bbox (M2B) conversion. The confidence of each box is calculated as the mean class probability of the highest-confidence 20\% of voxels within its component.

The complete inference procedure is summarized in Algorithm~\ref{alg:supp_inference}.

\section*{C. More Implementation Details}

\subsection*{C.1 Network Architecture}

AerialOcc adopts EfficientNet-B7 as the shared image backbone, followed by a feature pyramid network (FPN) to extract multi-scale visual features. Three adjacent RGB observations are processed by the shared encoder, and their features are aggregated through the Temporal Feature Fusion module.

For geometric modeling, the depth estimation network predicts monocular depth distributions and image context features, while a depth encoder extracts geometric features from projected LiDAR depth. A depth aggregation module combines these features to obtain the fused depth distribution. During training, a frozen VGGT model provides additional geometric supervision through KL-based distillation and is removed during inference.

The voxel encoder integrates temporal visual features and geometric cues using learnable voxel queries, LSS-generated 3D priors, and depth-guided attention. Local-global feature aggregation further refines the voxel representation. Finally, an occupancy prediction head maps the voxel features to semantic occupancy logits. Semantic segmentation and 3D object detection are derived from the shared occupancy prediction through voxel-wise classification and geometric post-processing, without additional learnable task-specific prediction heads.

\subsection*{C.2 Voxel Grid Configuration}

The original semantic occupancy annotations of AerialOcc-20K are generated on a \(256\times256\times192\) voxel grid. To match the model output resolution, we perform physical-coordinate 3D voxel resampling, mapping source voxel centers to the corresponding target voxels. The target grid has a resolution of \(256\times256\times32\), with voxel dimensions of \(0.30\times0.20\times1.25\,\mathrm{m}\) and a spatial range of
\([-25.6,51.2]\times[-25.6,25.6]\times[-32,8]\,\mathrm{m}\).

The original vertical coverage is preserved, while the horizontal X range is extended to accommodate the target grid. Semantic labels are aggregated from mapped source voxels, and validity masks are combined using an OR operation. Target voxels without valid source observations are excluded from occupancy supervision and evaluation.

The latent voxel grid has a resolution of \(128\times128\times16\), which is decoded into the final output grid of \(256\times256\times32\). Both occupancy prediction and evaluation are performed on this final grid.

\subsection*{C.3 Training Configuration}

AerialOcc is optimized for 25,000 iterations using AdamW, with a maximum learning rate of \(3\times10^{-4}\) and a weight decay of \(0.01\). We employ OneCycleLR with cosine annealing, where the first \(5\%\) of the optimization process, approximately 1,250 iterations, is used for learning-rate warm-up.

The batch size is set to 1 per GPU, resulting in a total batch size of 2 across two NVIDIA GeForce RTX 4090 GPUs. All experiments are conducted in FP32 without mixed-precision training, gradient clipping, or gradient accumulation.

The input images are resized to \(704\times384\).

\subsection*{C.4 Inference and Occupancy-Based Detection}

During inference, the frozen VGGT teacher is removed, and geometric features are obtained from the depth estimation and projected LiDAR branches. The model predicts semantic occupancy logits on the \(256\times256\times32\) grid, which are converted into voxel-wise class probabilities using softmax.

For 3D object detection, the predicted semantic labels are processed independently for each detectable category. We use 26-neighborhood connectivity for 3D connected-component analysis and remove components containing fewer than eight voxels. The retained components are converted into oriented 3D bounding boxes using M2B. The confidence of each box is determined by averaging the class probabilities of the highest-confidence 20\% of voxels within the corresponding component. All detection outputs are obtained from the shared semantic occupancy prediction without an additional learnable detection head.
\section*{D. More Experiments}
\subsection*{D.1 Per-Class Occupancy Results}
Table~\ref{tab:occ_perclass} reports the per-class semantic occupancy results of AerialOcc on AerialOcc-20K. AerialOcc achieves an overall IoU of 69.02\% and an mIoU of 29.93\%. It performs particularly well on major structural categories, including road, sidewalk, terrain, and ground, with IoUs above 70\%. Water, car, and rail track also achieve relatively strong results, with IoUs of 66.72\%, 50.77\%, and 57.09\%, respectively. In contrast, rare or small categories such as traffic light, pedestrian, truck, and guard rail remain challenging, while bus achieves only 0.25\% IoU. These results indicate that class imbalance and limited observations of small objects remain major challenges in fine-grained aerial occupancy prediction.

\begin{table*}[h]
\centering
\setlength{\tabcolsep}{12pt}
\begin{tabular}{lcc}
\toprule
\textbf{Category}
& \textbf{Distribution}
& \textbf{AerialOcc} \\
\midrule

\cbox{road} Road
& 30.13
& 87.92 \\

\cbox{sidewalk} Sidewalk
& 18.03
& 83.41 \\

\cbox{building} Building
& 22.57
& 38.07 \\

\cbox{wall} Wall
& 0.50
& 41.43 \\

\cbox{fence} Fence
& 0.34
& 16.22 \\

\cbox{pole} Pole
& 2.09
& 11.70 \\

\cbox{trafficlight} Traffic light
& 0.01
& 4.57 \\

\cbox{trafficsign} Traffic sign
& 0.04
& 10.96 \\

\cbox{vegetation} Vegetation
& 9.03
& 29.04 \\

\cbox{terrain} Terrain
& 8.20
& 76.42 \\

\cbox{pedestrian} Pedestrian
& 0.02
& 6.74 \\

\cbox{rider} Rider
& 0.01
& 4.45 \\

\cbox{car} Car
& 1.27
& 50.77 \\

\cbox{truck} Truck
& 0.52
& 5.96 \\

\cbox{bus} Bus
& 0.17
& 0.25 \\

\cbox{motorcycle} Motorcycle
& 0.01
& 12.20 \\

\cbox{bicycle} Bicycle
& 0.00
& 12.48 \\

\cbox{static} Static
& 0.38
& 20.12 \\

\cbox{dynamic} Dynamic
& 0.49
& 23.99 \\

\cbox{other} Other
& 0.34
& 23.71 \\

\cbox{water} Water
& 0.31
& 66.72 \\

\cbox{roadline} Road line
& 1.04
& 19.73 \\

\cbox{ground} Ground
& 0.55
& 72.88 \\

\cbox{bridge} Bridge
& 0.00
& 0.00 \\

\cbox{railtrack} Rail track
& 3.87
& 57.09 \\

\cbox{guardrail} Guard rail
& 0.04
& 1.31 \\

\midrule
\multicolumn{2}{l}{\textbf{Overall IoU}}
& \textbf{69.02} \\

\multicolumn{2}{l}{\textbf{mIoU}}
& \textbf{29.93} \\

\bottomrule
\end{tabular}
\caption{Per-class semantic scene completion results of AerialOcc on the AerialOcc-20K validation set.}
\label{tab:occ_perclass}
\end{table*}

\clearpage

\clearpage
\subsection*{D.2 Per-Class Results on OccuFly}
Table~\ref{tab:occufly_perclass} provides detailed results on OccuFly at different flight altitudes. Compared with CGFormer, AerialOcc improves occupancy IoU by 16.25, 17.08, and 23.82 percentage points at \(30\,\mathrm{m}\), \(40\,\mathrm{m}\), and \(50\,\mathrm{m}\), respectively. The corresponding mIoU improvements are 6.08, 2.54, and 6.31 percentage points. When all altitudes are combined, AerialOcc achieves 44.49\% IoU and 5.49\% mIoU, outperforming CGFormer by 19.83 and 4.69 percentage points. With only 10\% of the target-domain data for fine-tuning, AerialOcc obtains 5.44\% mIoU, only 0.20 percentage points below full-data fine-tuning. These results demonstrate consistent performance across altitude changes and effective transfer of semantic knowledge to real aerial environments.
\begin{table}[h]
\centering
\renewcommand{\arraystretch}{1.2}
\resizebox{\textwidth}{!}{%
\begin{tabular}{l|cc|cc|cc|cc|cc}
\hline
\multirow{2}{*}{\textbf{Category}}
& \multicolumn{2}{c|}{\textbf{30 m}}
& \multicolumn{2}{c|}{\textbf{40 m}}
& \multicolumn{2}{c|}{\textbf{50 m}}
& \multicolumn{2}{c|}{\textbf{All}}
& \multicolumn{2}{c}{\textbf{Sim-to-Real}} \\
\cline{2-11}
& \textbf{CGFormer}
& \textbf{AerialOcc}
& \textbf{CGFormer}
& \textbf{AerialOcc}
& \textbf{CGFormer}
& \textbf{AerialOcc}
& \textbf{CGFormer}
& \textbf{AerialOcc}
& \shortstack{\textbf{10\% FT}}
& \shortstack{\textbf{100\% FT}} \\
\hline
Road
& 3.87 & 11.10
& 2.21 & 2.42
& 0.28 & \textbf{20.09}
& 1.43 & 11.92
& 9.79 & \underline{12.23} \\
Walkway
& 1.22 & 2.07
& 0.55 & 0.17
& 0.55 & \textbf{6.39}
& 0.61 & 3.40
& 3.70 & \underline{3.82} \\
Dirt
& 0.17 & 1.24
& 0.09 & 0.22
& 0.31 & \textbf{4.23}
& 0.24 & 2.20
& \underline{2.67} & 2.32 \\
Gravel
& 0.16 & 1.20
& 0.23 & 0.30
& 0.44 & \underline{2.85}
& 0.35 & 1.71
& \textbf{2.97} & 2.09 \\
Rock
& 0.01 & 0.03
& 0.02 & 0.01
& \textbf{0.05} & 0.03
& \underline{0.04} & 0.02
& 0.01 & 0.03 \\
Grass
& 3.46 & \textbf{17.86}
& 1.03 & 3.81
& 8.51 & \underline{17.81}
& 5.35 & 11.53
& 15.36 & 14.58 \\
Vegetation
& 0.47 & \textbf{10.78}
& 0.35 & 2.62
& 0.92 & 4.18
& 0.76 & 4.07
& 4.30 & \underline{4.78} \\
Tree
& 0.54 & \textbf{20.12}
& 0.24 & 9.41
& 0.28 & 8.89
& 0.28 & 10.69
& \underline{11.78} & 10.35 \\
Ground-Obs.
& 0.98 & 1.43
& 0.09 & 0.24
& 0.41 & \textbf{6.58}
& 0.29 & 3.45
& 3.08 & \underline{4.16} \\
Person
& 0.00 & 0.00
& 0.00 & 0.00
& 0.00 & 0.00
& 0.00 & 0.00
& 0.00 & 0.00 \\
Bicycle
& 0.00 & 0.00
& 0.00 & 0.00
& 0.00 & 0.00
& 0.00 & 0.00
& \underline{0.03} & \textbf{0.04} \\
Vehicle
& 0.25 & \underline{14.21}
& 0.18 & 4.27
& 0.02 & \textbf{16.96}
& 0.11 & 10.99
& 13.09 & 11.66 \\
Water
& 0.00 & 0.00
& 0.00 & 0.00
& 0.00 & 0.00
& 0.00 & 0.00
& 0.00 & 0.00 \\
Building
& 9.59 & \underline{42.47}
& 7.67 & 34.89
& 7.90 & \textbf{46.37}
& 7.88 & 41.30
& 31.26 & 38.18 \\
Roof
& 0.68 & \textbf{10.52}
& 0.17 & 2.87
& 0.16 & \underline{7.83}
& 0.19 & 5.94
& 6.17 & 6.16 \\
Cables
& 0.00 & 0.00
& 0.00 & 0.00
& 0.00 & 0.00
& 0.00 & 0.00
& \textbf{0.01} & 0.00 \\
Cable-Tower
& 0.00 & 0.00
& 0.00 & 0.00
& 0.00 & 0.00
& 0.00 & 0.00
& 0.00 & 0.00 \\
Flying-Animals
& 0.00 & 0.00
& 0.00 & 0.00
& 0.00 & 0.00
& 0.00 & 0.00
& 0.00 & 0.00 \\
Parking-Lot
& 0.01 & \textbf{17.01}
& 0.00 & 5.29
& 0.01 & \underline{12.99}
& 0.01 & 10.22
& 10.51 & 9.49 \\
Constructions
& 0.05 & 0.71
& 0.06 & 0.04
& 0.15 & \textbf{0.98}
& 0.13 & 0.59
& 0.29 & \underline{0.78} \\
Cranes
& 0.00 & 0.00
& 0.00 & 0.00
& 0.00 & 0.00
& 0.00 & 0.00
& 0.00 & 0.00 \\
Truck
& 0.00 & \underline{4.28}
& 0.00 & 2.26
& 0.00 & 2.62
& 0.00 & 2.84
& \textbf{4.76} & 3.72 \\
\hline
\textbf{mIoU}
& 0.97 & \underline{7.05}
& 0.59 & 3.13
& 0.91 & \textbf{7.22}
& 0.80 & 5.49
& 5.44 & 5.64 \\
\textbf{IoU}
& 29.48 & \underline{45.73}
& 20.82 & 37.90
& 26.00 & \textbf{49.82}
& 24.66 & 44.49
& 33.39 & 41.18 \\
\hline
\end{tabular}%
}
\caption{Per-class semantic scene completion IoU results on the OccuFly validation set at different flight altitudes and adaptation settings. All results are reported in percentage form. FT denotes fine-tuning.}
\label{tab:occufly_perclass}
\end{table}

\clearpage
\section*{E. More Qualitative Analysis}
\subsection*{E.1 Dataset}

\begin{figure*}[h]
\centering
\includegraphics[width=0.78\textwidth, height=0.72\textheight, keepaspectratio=false]{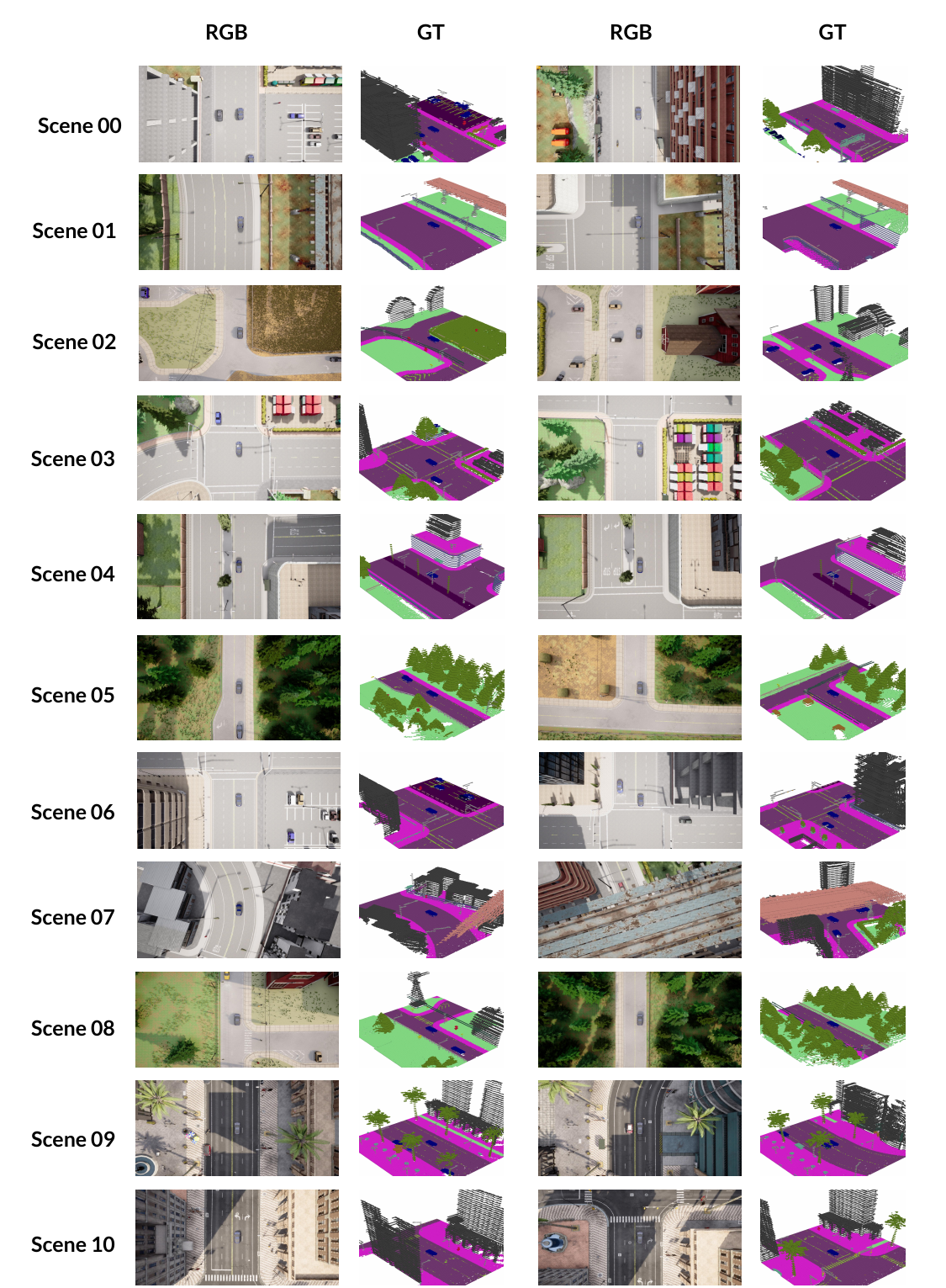}
\caption{Qualitative examples of RGB observations and the corresponding 3D semantic occupancy ground-truth annotations from scenes 00--10 in AerialOcc-20K.}
\label{fig:SupplementaryGT1}
\end{figure*}

\begin{figure*}[h]
\centering
\includegraphics[width=0.78\textwidth, height=0.72\textheight, keepaspectratio=false]{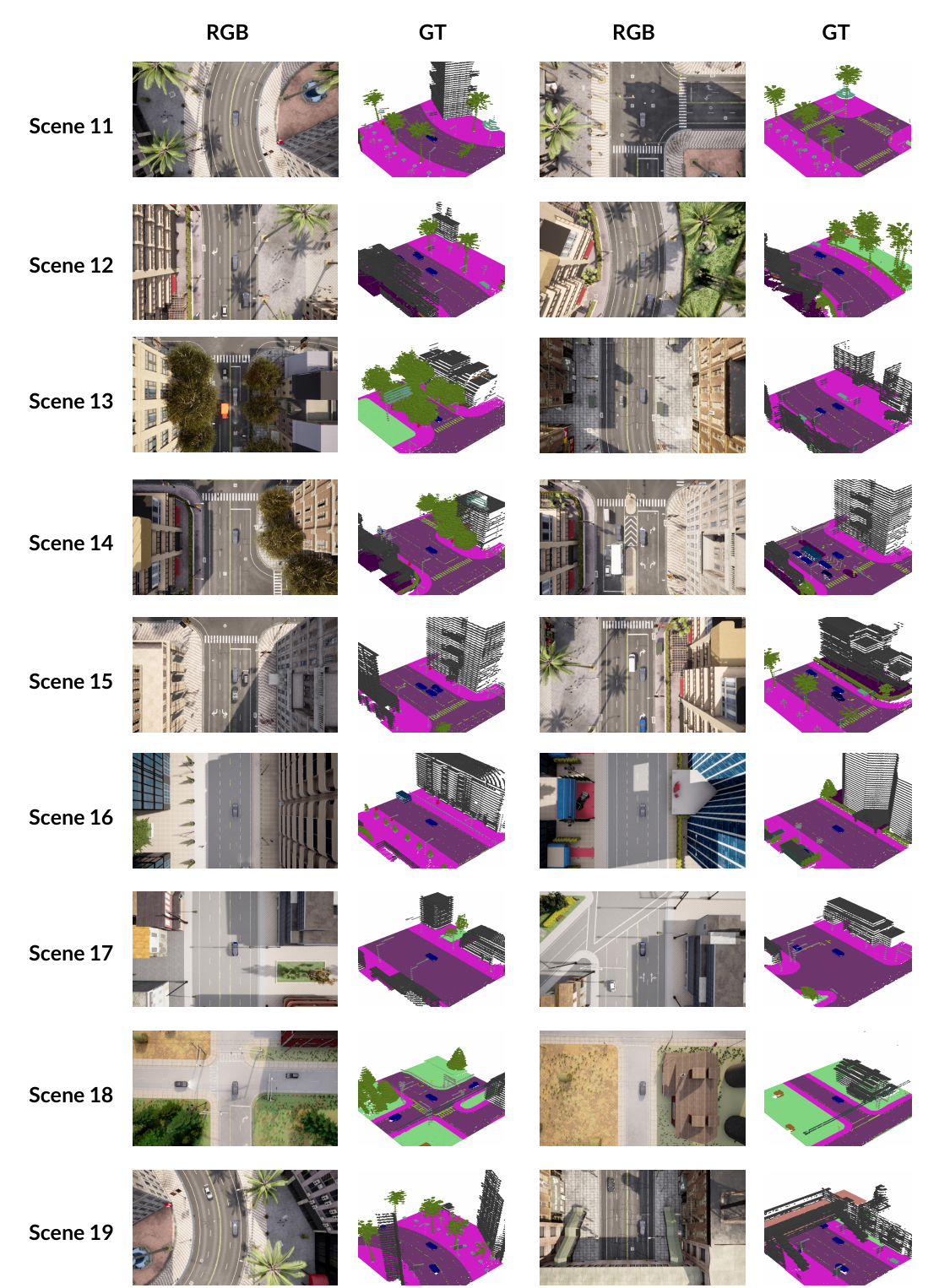}
\caption{Qualitative examples of RGB observations and the corresponding 3D semantic occupancy ground-truth annotations from scenes 11--19 in AerialOcc-20K.}
\label{fig:SupplementaryGT2}
\end{figure*}

\clearpage
\subsection*{E.2 Qualitative results of AerialOcc}

\begin{figure*}[h]
\centering
\includegraphics[width=1\textwidth]{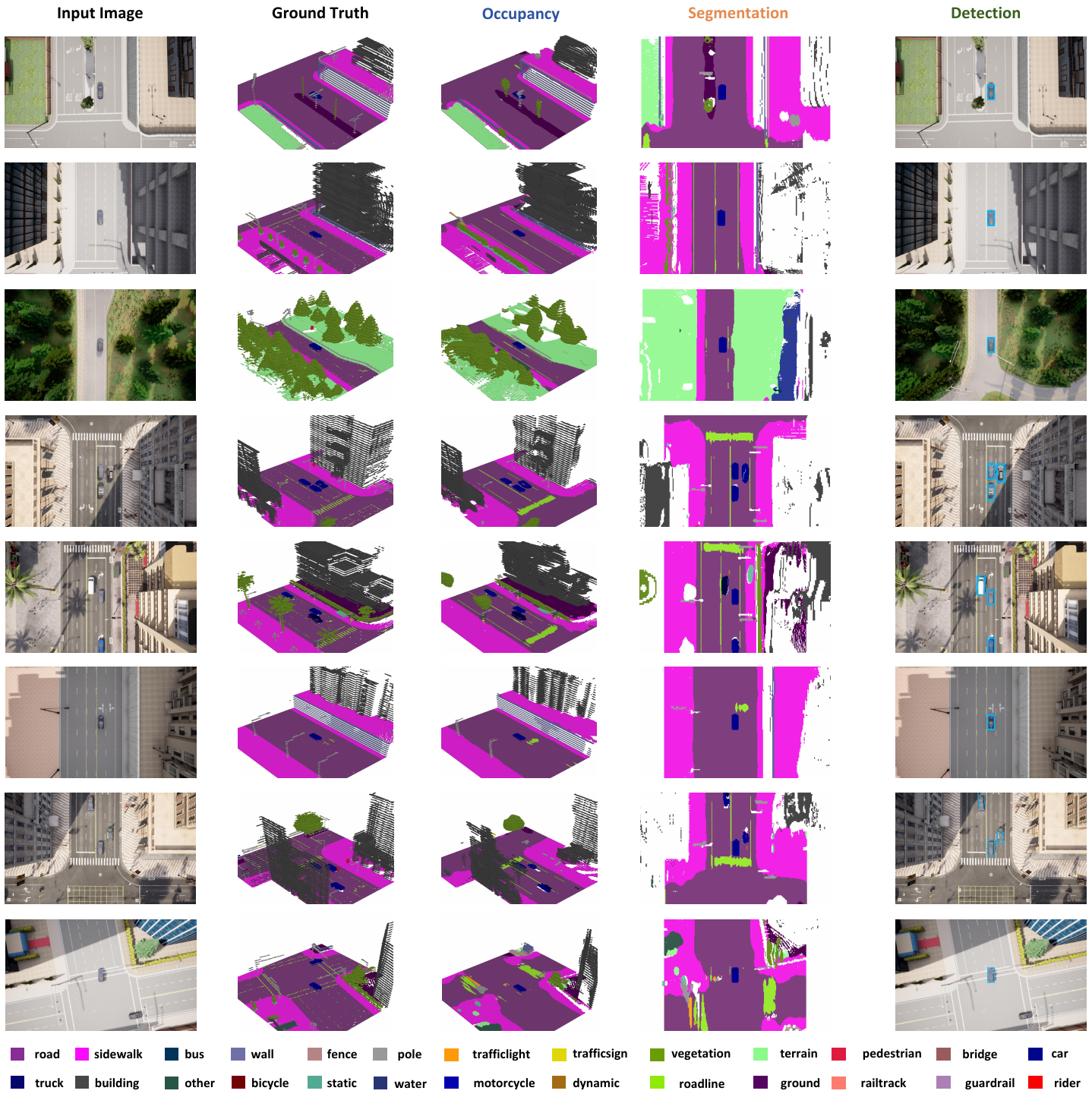} 
\caption{More qualitative results of AerialOcc on semantic scene completion, 3D voxel semantic segmentation, and 3D object detection.}
\label{fig:SupplementaryVisualization}
\end{figure*}


\clearpage
\section*{F. Limitations and Future Work}
AerialOcc currently adopts a fixed-resolution dense voxel grid to provide a consistent representation across different perception tasks. Although this design facilitates unified prediction and task conversion, its computational cost increases with the spatial range and voxel resolution. Existing sparse, hierarchical, or adaptive voxel representations could be incorporated to improve efficiency without changing the occupancy-centric formulation.

The current framework focuses on semantic scene completion, 3D voxel semantic segmentation, and 3D object detection. Future work may enrich the occupancy representation with additional attributes, such as motion, uncertainty, or temporal association, to support tasks including tracking and forecasting. These extensions are compatible with the unified occupancy interface and do not require redesigning the core framework.

\section*{G. Ethical Statement}
AerialOcc-20K is generated entirely in CARLA and therefore does not contain real-world personal information or identifiable individuals. The real-world benchmarks used in our experiments are employed only for research evaluation under their respective licenses.

Aerial perception technologies may support beneficial applications such as inspection, environmental monitoring, and autonomous navigation, but could also be misused for intrusive surveillance or unsafe autonomous operation. Practical deployment should comply with applicable privacy and aviation regulations and include appropriate safety constraints, human oversight, and access control.